\documentclass[10pt,twocolumn,letterpaper]{article}

\usepackage{cvpr}              % To produce the CAMERA-READY version
\definecolor{cvprblue}{rgb}{0.21,0.49,0.74}
\usepackage[pagebackref,breaklinks,colorlinks,allcolors=cvprblue]{hyperref}
\usepackage{booktabs}
\usepackage{multirow}
\usepackage{graphicx}
\usepackage{threeparttable}
\usepackage{tikz}
\usepackage[linesnumbered,ruled,vlined]{algorithm2e}
\SetKw{KwTo}{to}
\SetKw{KwEnd}{end}
\SetKw{KwFor}{for}
\usepackage{subcaption}
\usepackage{xcolor}
\usepackage{listings}
\usepackage{tabularx}
\usepackage{makecell}
\usepackage{bm}
\usepackage[table]{xcolor}
\usepackage{extpfeil}
\usepackage{array}

\def\paperID{xxxx} % *** Enter the Paper ID here
\def\confName{CVPR}
\def\confYear{2026}

\title{GroundVTS: Visual Token Sampling in Multimodal Large Language Models for Video Temporal Grounding}

\author{
Rong Fan$^{1,2}$\thanks{Equal contribution.} \quad Kaiyan Xiao$^{3}$\footnotemark[1] \quad Minghao Zhu$^{3}$ \quad  Liuyi Wang$^{3}$ \quad Kai Dai$^{1}$ \quad Zhao Yang$^{1}$\thanks{Corresponding author.}\\
{ $^{1}$Newcapec AI Research \quad $^{2}$Fudan University \quad $^{3}$Tongji University}\\
\tt \small rfan24@m.fudan.edu.cn \enspace \{xiaokaiyan,zmhh\_h,wly\}@tongji.edu.cn \enspace \{daikai,yangzhao\}@newcapec.net\\
}

\begin{document}
\maketitle

\begin{abstract}
Video temporal grounding (VTG) is a critical task in video understanding and a key capability for extending video large language models (Vid-LLMs) to broader applications.
However, existing Vid-LLMs rely on uniform frame sampling to extract video information, resulting in a sparse distribution of key frames and the loss of crucial temporal cues.
To address this limitation, we propose Grounded Visual Token Sampling (GroundVTS), a Vid-LLM architecture that focuses on the most informative temporal segments.
GroundVTS employs a fine-grained, query-guided mechanism to filter visual tokens before feeding them into the LLM, thereby preserving essential spatio-temporal information and maintaining temporal coherence.
Futhermore, we introduce a progressive optimization strategy that enables the LLM to effectively adapt to the non-uniform distribution of visual features, enhancing its ability to model temporal dependencies and achieve precise video localization.
We comprehensively evaluate GroundVTS on three standard VTG benchmarks, where it outperforms existing methods, achieving a 7.7-point improvement in mIoU for moment retrieval and 12.0-point improvement in mAP for highlight detection.
% Code is available at \url{https://github.com/Florence365/GroundVTS}.
Code is available at the \href{https://github.com/Florence365/GroundVTS}{GroundVTS code repository}.

\end{abstract}    
\section{Introduction}
\label{sec:intro}

\begin{figure*}[t]
    \centering
    % 第一张图
    \begin{subfigure}{0.33\linewidth}
        \centering
        \includegraphics[width=\linewidth]{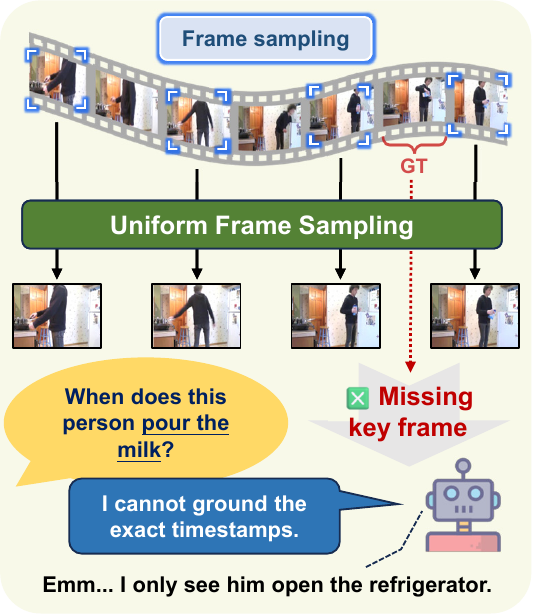}
        \caption{Uniform frame sampling.}
        \label{fig:uniform}
    \end{subfigure}%
    \hfill
    % 中间竖线
    % \vrule width 0.5pt height 5cm
    \raisebox{0.55cm}{\vrule width 0.5pt height 6.4cm}%
    \hfill
    % 第二张图
    \begin{subfigure}{0.33\linewidth}
        \centering
        \includegraphics[width=\linewidth]{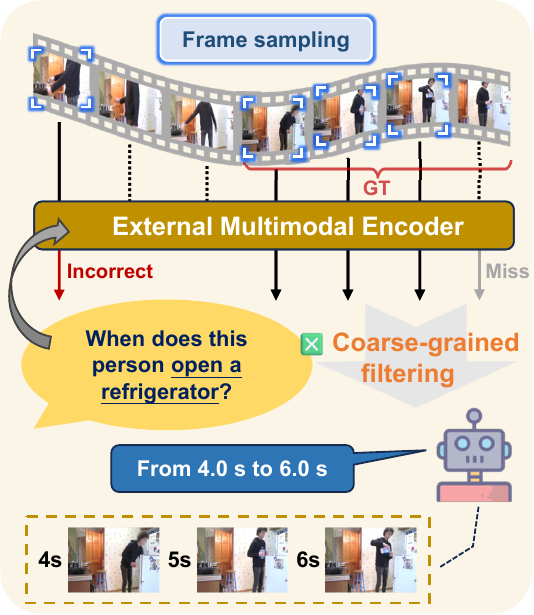}
        \caption{Query-guided frame sampling.}
        \label{fig:query}
    \end{subfigure}%
    \hfill
    % 中间竖线
    % \vrule width 0.5pt height 5cm
    \raisebox{0.55cm}{\vrule width 0.5pt height 6.4cm}%
    \hfill
    % 第三张图
    \begin{subfigure}{0.33
    \linewidth}
        \centering
        \includegraphics[width=\linewidth]{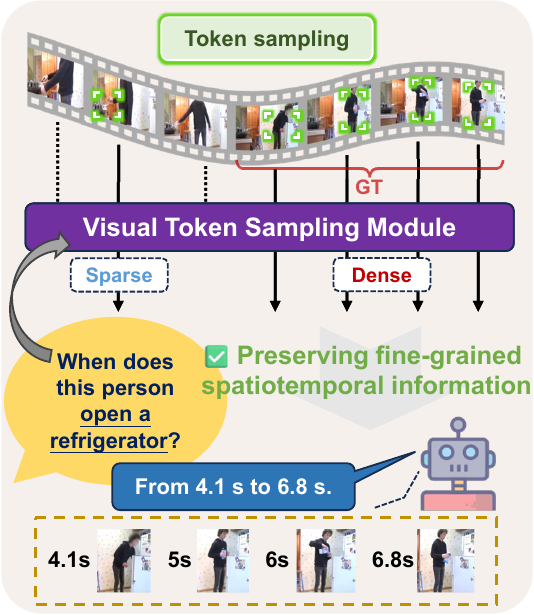}
        \caption{Ours: Query-guided visual token sampling.}
        \label{fig:ours}
    \end{subfigure}
    
    \caption{
        \textbf{Comparison of sampling strategies for Vid-LLMs.} 
        (a) Uniform sampling distributes attention evenly across frames, often missing query-relevant moments; (b) existing query-guided frame sampling relies on external encoders for coarse frame selection, limiting temporal understanding; and (c) our query-guided VTS adaptively selects query-relevant visual tokens within Vid-LLMs while maintaining temporal coherence, enabling efficient and precise video grounding.
    }
    \label{fig:architecture_fig}
\end{figure*}

Understanding complex video content is fundamental to a wide range of applications, including video classification~\cite{caba2015activitynet, goyal2017something, zhao2019hacs, han2024video}, video captioning~\cite{li2016tgif, zhou2018towards, chen2023vast}, video question-answering (VQA)~\cite{zellers2019recognition, yu2019activitynet, xiao2021next}, and video temporal grounding (VTG)~\cite{gao2017tall, krishna2017dense, lei2021detecting}.
Among these applications, VTG is a representative task for fine-grained temporal understanding: it aims to precisely identify the video segment corresponding to a given natural-language query~\cite{gao2017tall}.
%Among these, VTG serves as a key subtask of fine-grained temporal understanding, which requires precisely identifying the video segment corresponding to a given natural-language query~\cite{gao2017tall}.

Driven by the success of large-scale multimodal pretraining, numerous vision language models have achieved remarkable progress in multimodal reasoning~\cite{radford2021learning, sun2019videobert, zhu2020actbert, girdhar2023omnimae, tong2022videomae}.
Meanwhile, rapid advances in large language models (LLMs) have opened new opportunities to incorporate high-level reasoning and instruction following into video understanding.
When combined with visual encoders, LLMs provide a unified multimodal framework for interpreting complex temporal events and cross-modal relationships~\cite{tang2025video}.
Building on this trend, many studies have developed video LLMs (Vid-LLMs)~\cite{lin2023video, li2023videochat, maaz2023video}, achieving notable progress in general video reasoning.
However, current Vid-LLMs still struggle with granular temporal understanding.

Prior efforts to enhance Vid-LLMs for temporal grounding have introduced query-conditioned attention~\cite{zeng2024timesuite}, temporal boundary regressors~\cite{liu2025videomind}, and temporal modeling modules~\cite{bai2025qwen2}.
%Yet these approaches often overlook how visual tokens are sampled and represented before entering the LLM---a factor critical for precise temporal reasoning.
Yet these approaches often overlook the sampling and representation of visual tokens before entering the LLM, which can be important for precise temporal reasoning.
As shown in Figure~\ref{fig:architecture_fig}\subref{fig:uniform}, most existing approaches adopt uniform frame sampling, a dominant strategy in video understanding tasks.
While this strategy provides consistent temporal coverage, it allocates the input budget evenly across time; consequently, key moments can be diluted or even missed, especially when query-relevant events are sparse.
Recently, several studies~\cite{wang2025videotree, liang2024keyvideollm, wang2024videoagent} introduce query-guided frame sampling at the video input stage by attaching an external multimodal encoder (\emph{e.g.}, CLIP~\cite{radford2021learning} or a captioning model) to compute cross-modal similarity for frame selection (Figure~\ref{fig:architecture_fig}\subref{fig:query}).
However, they perform coarse-grained filtering and depend on auxiliary encoders, limiting localization precision and adaptability for VTG.

To address these limitations, we propose GroundVTS, a Vid-LLM architecture that performs query-guided visual token sampling at a finer granularity.
As illustrated in Figure~\ref{fig:architecture_fig}\subref{fig:ours}, GroundVTS introduces a Visual Token Sampling (VTS) module that operates after the visual encoder and multimodal projection layers.
Unlike the coarse selection used in prior work, VTS selectively retains visual tokens based on token-level similarity to the textual query, yielding a non-uniform token distribution across both temporal and spatial dimensions.
This helps the Vid-LLM focus on crucial cues while suppressing irrelevant visual content, enabling more precise and robust temporal grounding.

Our main contributions are summarized as follows:
\begin{itemize}
\item[$\bullet$] We propose GroundVTS, a Vid-LLM framework with a query-guided VTS strategy that adaptively preserves critical spatio-temporal cues to sharpen temporal localization for VTG.
%By focusing attention on query-relevant regions, VTS enables more precise, fine-grained video understanding.
% The VTS mechanism dynamically focuses the the model’s attention on query-relevant visual regions, leading to more precise temporal grounding.
\item[$\bullet$] We introduce a progressive optimization strategy that integrates the VTS module into existing Vid-LLM architectures, enabling effective adaptation to non-uniform visual token distributions while ensuring stable training.
%\item[$\bullet$] We introduce a \textbf{progressive optimization} strategy to integrate VTS into existing Vid-LLM architectures, enabling effective adaptation to non-uniform visual token distributions while maintaining stable training.
\item[$\bullet$] We conduct extensive experiments across multiple VTG benchmarks under varying visual token densities and model architectures, demonstrating the effectiveness and generality of GroundVTS.
\end{itemize}

% \begin{equation}
%   E = m\cdot c^2
%   \label{eq:important}
% \end{equation}
% and
% \begin{equation}
%   v = a\cdot t.
%   \label{eq:also-important}
% \end{equation}

% \url{http://www.pamitc.org/documents/mermin.pdf}.

% \begin{figure}[t]
%   \centering
%   \fbox{\rule{0pt}{2in} \rule{0.9\linewidth}{0pt}}
%    %\includegraphics[width=0.8\linewidth]{egfigure.eps}

%    \caption{Example of caption.
%    It is set in Roman so that mathematics (always set in Roman: $B \sin A = A \sin B$) may be included without an ugly clash.}
%    \label{fig:onecol}
% \end{figure}

% \begin{figure*}
%   \centering
%   \begin{subfigure}{0.68\linewidth}
%     \fbox{\rule{0pt}{2in} \rule{.9\linewidth}{0pt}}
%     \caption{An example of a subfigure.}
%     \label{fig:short-a}
%   \end{subfigure}
%   \hfill
%   \begin{subfigure}{0.28\linewidth}
%     \fbox{\rule{0pt}{2in} \rule{.9\linewidth}{0pt}}
%     \caption{Another example of a subfigure.}
%     \label{fig:short-b}
%   \end{subfigure}
%   \caption{Example of a short caption, which should be centered.}
%   \label{fig:short}
% \end{figure*}

\section{Related Work}
\label{sec:relate}

{\bf Video large language models (Vid-LLMs)} are typically implemented by connecting visual encoders to LLMs through projection layers or multimodal adapters.
Recent studies have focused on enhancing the global video understanding capabilities~\cite{lin2023video, li2023videochat, maaz2023video, wang2024internvideo2, wang2025videotree}.
While these models achieve strong performance, they still struggle to capture granular temporal structures~\cite{liu2024tempcompass}.
GroundVTS instead emphasizes fine-grained temporal perception, thereby improving the ability of Vid-LLMs to handle challenges of VTG.

\noindent{\bf Video temporal grounding (VTG)} is a fundamental video understanding task that aims to accurately localize the timestamps of events within a video~\cite{lin2023univtg}.
Classical expert models formulate this as a cross-modal matching problem, employing proposal-based or regression-based architectures to improve grounding precision~\cite{liu2021progressively, hu2024maskable, yang2024task, zeng2024unimd}.
%While these models achieved early success, they were typically designed for single-task scenarios, and required extensive fine-tuning for each downstream dataset~\cite{wang2024timerefine}.
These models are typically task-specific and require dataset-level fine-tuning~\cite{wang2024timerefine}.
Recent studies~\cite{wu2025number, deng2025seq2time, huang2024vtimellm, huang2024lita, qian2024momentor, guo2025vtg} integrate VTG capabilities into Vid-LLMs, leveraging their strong reasoning and generalization strengths to unify diverse video understanding tasks within a single framework.
%For instance, VTG-LLM~\cite{guo2025vtg} integrates specialized time tokens and temporal position embeddings to enhance Vid-LLMs' understanding of timestamps, while NumPro~\cite{wu2025number} represents video frames as numbered image sequences to strengthen the link between visual perception and temporal understanding.
To support VTG within this unified setting, many methods introduce explicit time representations (\emph{e.g.}, time tokens/ position embeddings) or frame indexing to better model timestamps~\cite{guo2025vtg, wu2025number}.
However, these approaches primarily focus on temporal encoding or data adaptation; the sampling and representation of visual tokens before entering the LLM have received comparatively less attention.
GroundVTS addresses this gap by introducing a unified Vid-LLM framework with an effective token sampling mechanism that substantially improves temporal understanding.

\noindent{\bf Visual token compression and selection.}
In VTG and VQA, not all visual tokens contribute equally to answering a given query, and many contain redundant or irrelevant information that can distract the model from the relevant evidence~\cite{park2024too}. 
This has motivated growing interest in token compression, pruning, and adaptive selection for Vid-LLMs~\cite{alvar2025divprune, tan2025tokencarve, huang2024ivtp}.
However, most existing approaches rely on query-agnostic or saliency-based token reduction, often preserving visually prominent yet semantically irrelevant tokens, which can limit temporal reasoning and grounding precision.
Recent token compression methods further deliver strong efficiency gains and improve scalability for long-video understanding~\cite{zhang2024sparsevlm, li2024llamavid, liu2025hybrid}.
Their implications for VTG---where fine-grained temporal localization is central---remain less explored.
In contrast, our GroundVTS dynamically samples query-relevant tokens within the Vid-LLM via token--query relevance, preserving spatio-temporal cues without external preprocessing.
\section{Method}
\label{sec:methods}
We propose GroundVTS, a Vid-LLM architecture designed to enhance VTG performance through adaptive and efficient visual token utilization. 
Sec.~\ref{sec:preliminary_study} analyzes the sensitivity of VTG performance to frame density, which highlights the necessity of adaptive token sampling, motivating our design.
Sec.~\ref{sec:architecture} presents the overall architecture of GroundVTS, followed by a detailed explanation of the Visual Token Sampling (VTS) module in Sec.~\ref{sec:vts_module}.
Finally, Sec.~\ref{sec:train_stages} outlines our progressive optimization strategy, which allows VTS to be seamlessly integrated into existing Vid-LLMs.

%-------------------------------------------------------------------------
%-------------------------------------------------------------------------
\begin{figure}[!t]
  \centering
  \includegraphics[width=\linewidth]{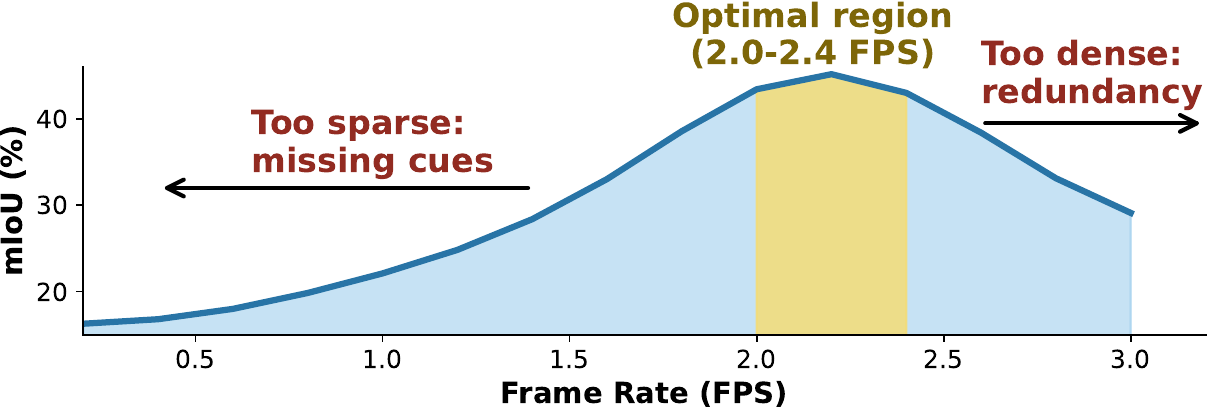}
  \caption{Frame rate sensitivity of Qwen2.5VL-7B. Similar trends hold for InternVL3.5 (illustrated in the supplementary material).}
  \label{fig:fps_sensitivity}
\end{figure}
%-------------------------------------------------------------------------
\subsection{Analysis of Frame Rate Sensitivity}\label{sec:preliminary_study}
Before introducing our sampling strategy, we first examine how the density of visual tokens inherently influences VTG performance.
To this end, we evaluate the pretrained Qwen2.5VL-7B~\cite{bai2025qwen2} model on the Charades-STA~\cite{gao2017tall} dataset under varying frame rates from 0.2 to 3.0 frames per second (FPS), while keeping all other settings fixed.

As shown in Figure~\ref{fig:fps_sensitivity}, VTG performance demonstrates a clear non-linear dependency on frame rate. 
When the frame rate is low ($<$1.0 FPS), the model lacks sufficient temporal cues, resulting in degraded mIoU.
As the sampling rate increases, performance improves steadily and reaches its peak around 2.0--2.4 FPS, where mIoU attains 47.8\%. 
However, further increasing the frame rate beyond this range leads to a sharp performance drop, indicating that redundant visual tokens dilute key temporal signals and hinder accuracy.

This observation supports our core hypothesis that the density and relevance of visual tokens critically influence VTG performance. 
Consequently, an adaptive token sampling mechanism is essential for achieving both accuracy and efficiency in temporal grounding, motivating the design of our GroundVTS framework.

%-------------------------------------------------------------------------
\subsection{GroundVTS}\label{sec:architecture}
\begin{figure*}
  \centering
  \begin{subfigure}{0.57\linewidth}
    \includegraphics[width=\linewidth]{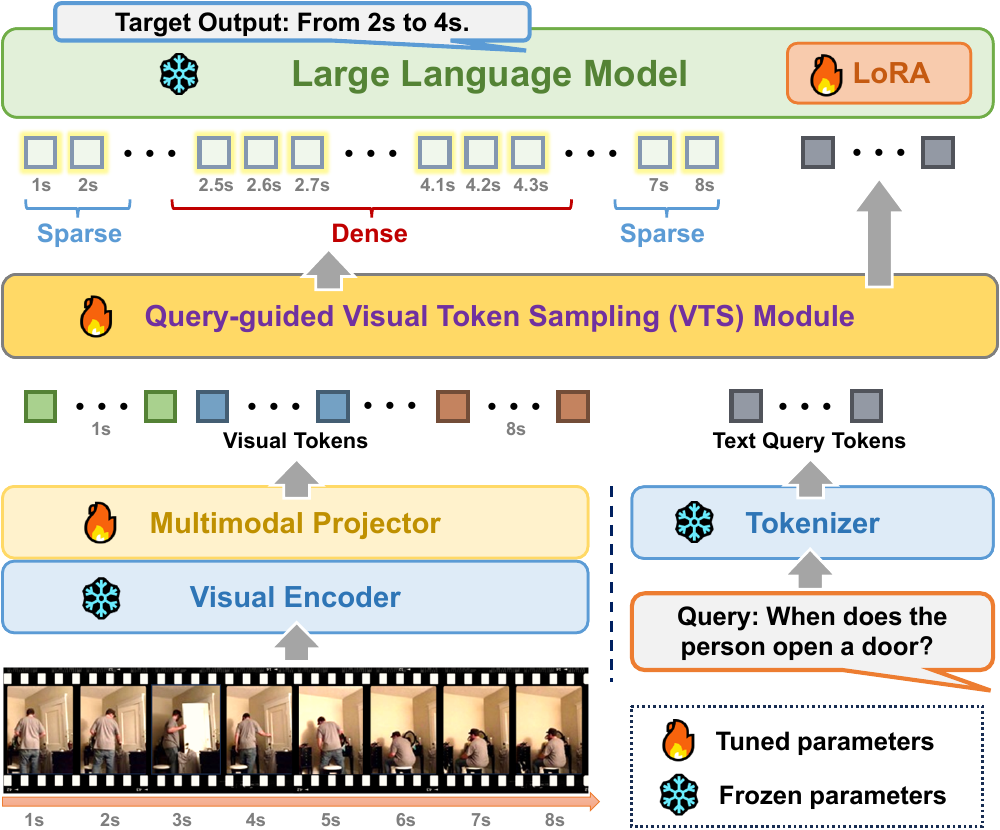}
    \caption{Overall architecture.}
    \label{fig:overall}
  \end{subfigure} \quad
  \begin{subfigure}{0.38\linewidth}
    \includegraphics[width=\linewidth]{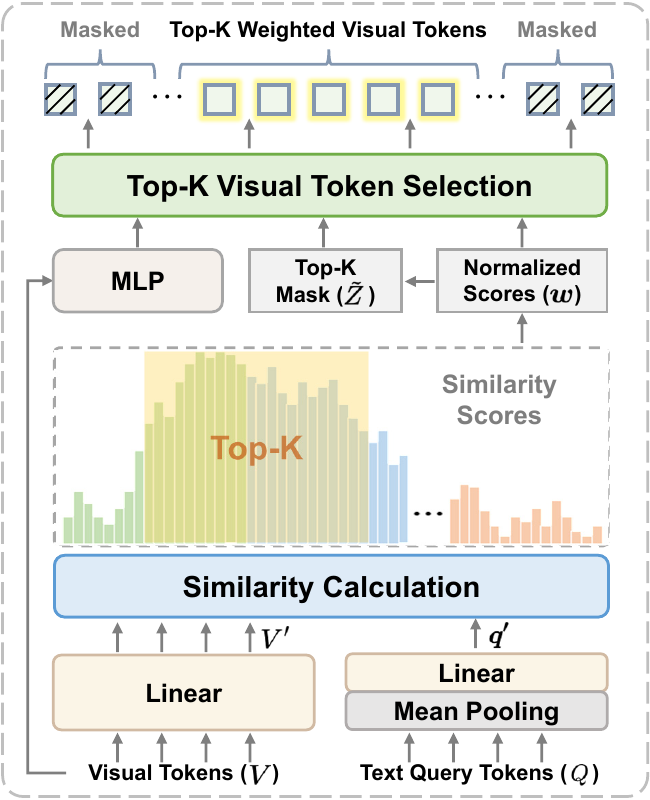}
    \caption{The VTS module.}
    \label{fig:vts}
  \end{subfigure}
  \caption{
    \textbf{Overview of the proposed GroundVTS framework.} 
    (a) GroundVTS integrates a query-guided VTS module into the Vid-LLM pipeline, enabling adaptive selection of query-relevant tokens;  
    (b) The VTS module computes token-query similarity scores and performs weighted differentiable top-$K$ sampling to retain the most informative tokens, supporting efficient and precise video temporal grounding.
  }
  \label{fig:architecture}
\end{figure*}

The GroundVTS framework, illustrated in Figure~\ref{fig:architecture}\subref{fig:overall}, incorporates a query-guided visual token sampling module to enable efficient and fine-grained temporal grounding.
Unlike prior methods that rely on additional preprocessing~\cite{wang2025videotree}, 
GroundVTS integrates the sampling process directly into the Vid-LLM pipeline, allowing the model to dynamically focus on temporally and semantically relevant segments conditioned on the input query.

Specifically, given an input video after standard temporal downsampling, $\mathcal{V}=\{F_t\}_{t=1}^T$, and a text query, the query is first tokenized into a sequence of embeddings, $Q=[\bm{q_1}, \bm{q_2}, \ldots, \bm{q_{N_t}}] \in \mathbb{R}^{N_t\times D}$, using a pretrained language tokenizer.
Here, $T$ denotes the number of frames, $N_t$ is the number of text tokens, and $D$ is the text token embedding dimension.
Each video frame $F_t$ is divided into spatial patches and encoded by a pretrained vision encoder to produce dense spatio-temporal features, $H_v=[\bm{h_1},\bm{h_2},\ldots,\bm{h_{N_v}}]\in\mathbb{R}^{N_v\times D_v}$, where $N_v$ denotes the total number of visual tokens and $D_v$ denotes the feature dimension.
Then, a multimodal projector (implemented as a multilayer perceptron) maps these features to a shared embedding space: $V=[\bm{v_1},\bm{v_2},\ldots,\bm{v_{N_v}}]\in\mathbb{R}^{N_v\times D}$, where each $\bm{v_i} \in \mathbb{R}^D$ is the projected embedding of the $i$-th visual token, ensuring that visual embeddings are dimensionally aligned with the text token embeddings $Q$.

Conditioned on the query embeddings $Q$, the VTS module (which will be introduced in Sec.~\ref{sec:vts_module}) evaluates the relevance of each visual token to the query and outputs a compact subset of visual token embeddings via a weighted differentiable top-$K$ selection mechanism:
\begin{equation}
\widetilde{V}=\operatorname{VTS}(V,Q)=[\bm{\tilde{v}_{1}}, \bm{\tilde{v}_{2}}, \ldots,\bm{\tilde{v}_{K}}],
\label{eq:vts_sampling}
\end{equation}
where $K$ is the number of tokens selected by VTS.
The resulting $\widetilde{V}$ forms a non-uniform, query-guided distribution of visual tokens that allocates denser sampling around relevant moments and sparser coverage elsewhere.

To preserve temporal coherence under non-uniform sampling, we reuse the original positional encodings from dense sampling (masking out only those of unselected visual tokens), ensuring that the selected tokens remain temporally aligned with the input video.

Finally, the sampled visual tokens and query embeddings are concatenated to form a multimodal sequence and processed by the LLM for joint reasoning and generation.
%to form a multimodal sequence, $X = [Q, \widetilde{V}]$, which is

%-------------------------------------------------------------------------
\subsection{Visual Token Sampling}\label{sec:vts_module}
The VTS module is the core of GroundVTS, responsible for dynamically selecting the most informative visual tokens under the guidance of the textual query.
As illustrated in Figure~\ref{fig:architecture}\subref{fig:vts}, VTS consists of the following two main steps.

{\bf Query-Guided Token Scoring.}
Given the projected visual embeddings, $V=\{\bm{v_i}\}_{i=1}^{N_v}$, and the tokenized query embeddings, $Q=\{\bm{q_j}\}_{j=1}^{N_t}$, to estimate the relevance between each visual token and the query, both sets of embeddings are first projected into lower-dimensional subspaces of dimension $D_r$ through linear projections:
\begin{equation}
V'=W_vV,\ \bm{q'}=W_q\mathrm{Pool}\left(Q \right),\ W_v,W_q\in\mathbb{R}^{D\times D_r}.
\end{equation}
Here, $W_v$ and $W_q$ are trainable projection matrices, and $\mathrm{Pool}(\cdot)$ denotes mean pooling over text token embeddings.

Then, the relevance distribution is obtained through a softmax function applied to the temperature-scaled dot products between each pair of embeddings in $V'$ and $\bm{q'}$:
\begin{equation}
\bm{w} = \operatorname{softmax}\left(V' \bm{q'}^{\top} / \tau \right),
\end{equation}
where $\tau$ is a temperature hyperparameter controlling the sharpness of the distribution.
This formulation can be interpreted as an attention mechanism, where the weight assigned to each visual token reflects both its alignment with the query and its relative importance within the sequence.
The resulting weight vector, $\bm{w} = [w_1, w_2, \dots, w_{N_v}]^\top$, effectively emphasizes semantically relevant visual tokens while attenuating less informative ones.

{\bf Differentiable Top-$\bm{K}$ Selection.}
To efficiently retain the most relevant visual information, we select the top-$K$ visual tokens based on $\bm{w}$.
The number of selected tokens is adaptively determined by the ratio $\rho \in (0,1]$, where $K = \lceil \rho \cdot N_v \rceil$.
Let $\mathcal{I}_K$ denote the indices of the top-$K$ tokens.

Since hard top-$K$ selection is non-differentiable, we employ a Straight-Through Estimator (STE) with Gumbel-Softmax relaxation~\cite{jang2016categorical} to enable end-to-end training.
Specifically, we add Gumbel noise $g_i \sim \text{Gumbel}(0,1)$ to the log-probabilities and compute a differentiable approximation of the discrete selection mask:
\begin{equation}
z_i =
\frac{\exp\left((\log w_i + g_i)/\tau_g \right)}{\sum_{j=1}^{N_v} \exp\left((\log w_j + g_j)/\tau_g \right)},
\end{equation}
where $\tau_g$ is a Gumbel temperature controlling the smoothness of the relaxation.
During the forward pass, a hard top-$K$ operator is applied:
\begin{equation}
z_i^{\text{hard}} = 
\begin{cases}
1, & \text{if} \ i \in \mathcal{I}_K, \\
0, & \text{otherwise},
\end{cases}
% \quad \widetilde{S} = \left\{ \widetilde{s}_i \right\}_{i=1}^{N_v}.
\end{equation}
while the backward propagation flows through the continuous relaxation $z_i$. To this end, the STE is implemented as the following, combining the hard and soft representations:
\begin{equation}
\tilde{z}_i = z_i^{\text{hard}} + z_i - \mathrm{stopgrad}(z_i),
\end{equation}
where $\mathrm{stopgrad}(\cdot)$ denotes gradient detachment.

The final visual representations are then obtained via weighted differentiable top-$K$ masking: the relevance weights of the selected top-$K$ tokens are re-normalized to sum to one and serve as the non-zero elements of the final mask, while all non-top-$K$ positions remain zero.
Mathematically, this is formulated as:
\begin{equation}
\bm{\tilde{v}}_i = \hat{w}_i \cdot \operatorname{MLP}(\bm{v}_i), \quad
\hat{w}_i = \frac{\exp\left(w_i/\tau'\right)\cdot \tilde{z}_i}{\sum_{j=1}^{N_v} \exp\left(w_j/\tau'\right)\cdot \tilde{z}_j},
\end{equation}
where $\operatorname{MLP}$ indicates a multilayer perceptron and $\tau'$ represents the temperature hyperparameter.
This strategy ensures that token selection remains query-aware and trainable.

% Specifically, the similarity scores are masked such that all tokens outside the Top-$K$ set are assigned a value of $-\infty$:
% \begin{equation}
% \widetilde{s}_i = 
% \begin{cases}
% s_i, & \text{if} \ i \in \mathcal{I}_K, \\
% -\infty, & \text{otherwise}.
% \end{cases},
% \quad \widetilde{S} = \left\{ \widetilde{s}_i \right\}_{i=1}^{N_v}.
% \end{equation}

% The masked similarity scores $\widetilde{S}$ are then converted into a soft importance distribution using the softmax function:
% \begin{equation}
% w_i = \frac{\exp(\widetilde{s}_i)}{\sum_{j=1}^{N_v} \exp(\widetilde{s}_j)}, \quad W=\{ w_i\}^{N_v}_1.
% \end{equation}
% The soft importance weights $W$ are used for backpropagation:
% \begin{equation}
% \widetilde{v}_i = 
% \begin{cases}
% \mathrm{MLP}(v'_i) \cdot w_i, & \text{if } i \in \mathcal{I}_K, \\
% 0, & \text{otherwise}.
% \end{cases}, \quad \widetilde{V}=\{ \widetilde{v} \}^{N_v}_1.
% \end{equation}
% In this manner, the initial dense visual embeddings $V'$ are augmented with the query-guided attention scores $W$, enhancing their representation. This approach ensures stable end-to-end optimization without introducing sampling noise during the forward pass, leading to efficient and precise learning.

%-------------------------------------------------------------------------
\subsection{Progressive Optimization Strategy}\label{sec:train_stages}
To ensure stable convergence and effective cross-modal adaptation, GroundVTS is optimized through a three-stage progressive training strategy.
Each stage focuses on a specific learning objective, allowing the model to gradually develop non-uniform visual token sampling and query-conditioned reasoning capabilities while maintaining the stability of the underlying Vid-LLM.

{\bf Stage 1: VTS Warm-up.}
The first stage aims to initialize the query-guided visual token sampling process.
Since the VTS module dynamically samples tokens based on query relevance, jointly training it with the LLM from scratch can lead to unstable gradients and inconsistent selection behavior.
To mitigate this, we train only the parameters of the VTS module while freezing all other components.
This warm-up phase enables VTS to learn robust visual token-query relevance estimation and stable visual importance prediction before engaging in joint optimization.

{\bf Stage 2: Joint LoRA Adaptation.}
After the VTS has learned stable sampling behavior, the second stage focuses on aligning the non-uniform visual token distributions with textual semantics.
We fine-tune the LLM using Low-Rank Adaptation (LoRA)~\cite{hu2022lora} while jointly updating the VTS and the multimodal projector.
This stage leverages a large-scale multimodal dataset LLaVA-Video-178K~\cite{zhang2024videoinstructiontuningsynthetic}, exposing the model to diverse temporal structures and cross-modal reasoning scenarios.
Through this joint optimization, the model learns to effectively interpret and reason over query-guided, non-uniform visual token sequences, enabling fine-grained and efficient temporal reasoning within the LLM.
% Through joint optimization, the model learns to interpret query-dependent token subsets effectively, enhancing the coupling between selected frames and linguistic cues.

{\bf Stage 3: Grounding Fine-tuning.}
The final stage aims to refine the model’s temporal grounding and reasoning ability for real-world VTG tasks. 
We fine-tune GroundVTS on a newly curated dataset, Grounding-FT (see Sec.~\ref{sec:exp_setup}), which aggregates training samples from multiple temporal grounding datasets. 
% Each instance in Grounding-FT is formatted as an instruction-style prompt, guiding the model to predict temporal boundaries or highlight relevant events in a question-answering format, consistent with the previous stages. 
Each instance in Grounding-FT follows the same instruction-style QA format as in previous stages, and is used to guide the model to predict temporal boundaries or highlight relevant events.
This unified instruction-based interface enables GroundVTS to perform temporal grounding and descriptive reasoning jointly within a single generative process. 
Throughout this stage, the module freezing configuration follows that of Stage 2, ensuring smooth and stable fine-tuning.

\section{Experiments}
\label{sec:experiments}

%-------------------------------------------------------------------------
\begin{table*}[!t]
\centering
\caption{Comparison with state-of-the-art methods on Charades-STA and ActivityNet-Captions test splits.}
\label{tab:main_1}
\resizebox{\textwidth}{!}{%
\begin{tabular}{l|*{4}{>{$}l<{$}}|*{4}{>{$}l<{$}}}
\toprule
\multirow{2}{*}{Method} & \multicolumn{4}{c|}{Charades-STA} & \multicolumn{4}{c}{ActivityNet-Captions} \\ 
                       & \text{R1@.3}  & \text{R1@.5}  & \text{R1@.7} & \text{mIoU} & \text{R1@.3}    & \text{R1@.5}   & \text{R1@.7}   & \text{mIoU}   \\ \midrule
LLaVA-OV\cite{li2024llava} {\color{gray}\small arXiv' 24} & 28.8 & 16.6 & 5.9  & 19.3 & 20.2 & 8.6  & 2.2 & 13.5 \\
TimeChat\cite{ren2024timechat} {\color{gray}\small CVPR' 24}        & 47.7  & 22.9  & 12.5  & 30.6  & 30.2  & 16.9  & 8.2  & 21.8  \\
VTimeLLM\cite{huang2024vtimellm} {\color{gray}\small CVPR' 24}       & 51.0    & 27.5  & 11.4  & 31.2  & 44.0    & 27.8  & 14.3 & 30.4  \\
Momentor\cite{qian2024momentor} {\color{gray}\small ICML' 24}       & 42.9  & 23.0    & 12.4  & 29.3  & 42.6  & 26.6  & 11.6 & 28.5  \\
HawkEye\cite{wang2024hawkeye} {\color{gray}\small arXiv' 24}        & 50.6  & 31.4  & 14.5  & 33.7  & 49.1  & 29.3  & 10.7 & 32.7  \\
ChatVTG\cite{qu2024chatvtg} {\color{gray}\small CVPR' 24}        & 52.7 & 33.0 & 15.9 & 34.9 & 40.7 & 22.5 & 9.4 & 27.2 \\
NumPro\cite{wu2025number} {\color{gray}\small CVPR' 25}         & \underline{63.8}  & 42.0    & 20.6  & 41.4  & \textbf{55.6}  & \textbf{37.5}  & \underline{20.6} & \textbf{38.8}  \\
LLaVA-ST\cite{li2025llava} {\color{gray}\small CVPR' 25}         & 63.1  & \underline{44.8}    & 23.4  & \underline{42.4}  &--   &--   &--  &--   \\  \midrule
Qwen2.5VL-7B      & 34.2	& 18.8	& 8.6	& 22.1	& 25.3	& 11.5	& 4.4	& 17.1
  \\ 
Qwen2.5VL-7B-G     & 45.2	& 32.7	& 18.7	& 31.7	& 40.6	& 23.9	& 9.9	& 26.7
  \\ 
\rowcolor{gray!15}
GroundVTS-Q~(ours)      & \textbf{71.5}_{\color{teal}(\uparrow 26.3)}  & \textbf{57.5}_{\color{teal}(\uparrow 24.8)}  & \textbf{34.2}_{\color{teal}(\uparrow 15.5)}  & \textbf{50.1}_{\color{teal}(\uparrow 18.4)}    & \underline{51.3}_{\color{teal}(\uparrow 10.7)}  & \underline{33.6}_{\color{teal}(\uparrow 9.7)}  & \textbf{21.4}_{\color{teal}(\uparrow 11.5)} & \underline{36.0}_{\color{teal}(\uparrow 9.3)}  \\ \bottomrule
InternVL3.5-8B      & 35.5	& 25.7	& 13.2	& 24.6	& 22.1	& 12.0	& 5.6  & 15.8	
  \\ 
InternVL3.5-8B-G      & 59.5	& 42.0	& 20.2	& 39.4	& 35.9	& 20.6	& 9.0	& 24.5
  \\ 
\rowcolor{gray!15}
GroundVTS-I~(ours)      & 61.2_{\color{teal}(\uparrow 1.7)}  & 44.2_{\color{teal}(\uparrow 2.2)}  & \underline{23.7}_{\color{teal}(\uparrow 3.5)}  & 41.6_{\color{teal}(\uparrow 2.2)}    & 37.9_{\color{teal}(\uparrow 2.0)}  & 22.4_{\color{teal}(\uparrow 1.8)}  & 10.3_{\color{teal}(\uparrow 1.3)} & 25.7_{\color{teal}(\uparrow 1.2)}  \\ \bottomrule
\end{tabular}%
}
\vspace{3pt}
\footnotesize{%
\begin{minipage}{\linewidth}
\raggedright
\textbf{Bold} denotes the best, \underline{underlined} denotes the second-best. ``-G'' denotes supervised fine-tuning on the Grounding-FT dataset. ``-Q'' and ``-I'' denote our proposed models based on Qwen2.5VL-7B~\cite{bai2025qwen2} and InternVL3.5-8B~\cite{wang2025internvl3}, respectively. {\color{teal}$\uparrow$} indicates improvement over the corresponding ``-G'' baseline.
\end{minipage}
}
\end{table*}

%$^{\dagger}$
%$^{\dagger}$
%$^{*}$
%$^{*}$
%-------------------------------------------------------------------------
\subsection{Experimental Setup}\label{sec:exp_setup}
{\bf Dataset Preparation.}
We employ two datasets in the training pipeline:
{\bf (a) LLaVA-Video-178K}~\cite{zhang2024videoinstructiontuningsynthetic}, a large-scale video dataset that provides diverse multimodal supervision across tasks such as video captioning and VQA for pretraining and alignment;
and {\bf (b) Grounding-FT}, a curated dataset we construct for VTG tasks. To adapt VTG supervision to the natural language input-output format of LLMs, we design a set of instruction templates with diverse linguistic expressions and combine them with temporal grounding queries to form QA-style training pairs. 
Grounding-FT is derived from multiple VTG training splits covering both moment retrieval and highlight detection tasks~\cite{gao2017tall, lei2021detecting, krishna2017dense}, and contains a total of 70K annotated video-query pairs.
Dataset details are provided in the supplementary material.
LLaVA-Video-178K is used in the first two training stages, while Grounding-FT is employed in the third stage.

\noindent{\bf Evaluation.}
We evaluate our model on two representative VTG tasks, namely moment retrieval (MR)~\cite{caba2015activitynet, gao2017tall} and highlight detection (HD)~\cite{lei2021detecting}.
The MR task aims to identify the start and end timestamps of the video segment corresponding to a given natural language query.
Following standard practice, we conduct evaluation on Charades-STA~\cite{gao2017tall}, ActivityNet-Captions~\cite{caba2015activitynet}, and QVHighlights~\cite{lei2021detecting}, using mean intersection-over-union (mIoU) and Recall$@1$ (R1$@t$) at thresholds $t \in$ \{0.3, 0.5, 0.7\}~\cite{qu2024chatvtg, wu2025number, li2025llava}.
The HD task requires the model to output all salient moments relevant to the query in the video together with their corresponding relevance scores. We use QVHighlights~\cite{lei2021detecting} for evaluation and adopt mean average precision (mAP) and the hit ratio of the highest-scored clip (Hit@1) as metrics~\cite{lin2023univtg, guo2025vtg, ren2024timechat}.

\noindent{\bf Implementation Details.}
We construct two model variants, GroundVTS-Q and GroundVTS-I, built upon Qwen2.5VL-7B~\cite{bai2025qwen2} and InternVL3.5-8B~\cite{wang2025internvl3}, respectively.
Both models are trained using the three-stage strategy described in Sec.~\ref{sec:train_stages}, where stages 1--3 are trained for 1, 2, and 3 epochs, respectively, with learning rates of $1\!\times\!10^{-5}$, $2\!\times\!10^{-4}$, and $1\!\times\!10^{-4}$.
The two base models differ in their intrinsic video sampling paradigms.
QwenVL employs a fixed frame-rate strategy, uniformly sampling frames over time, whereas InternVL adopts a fixed frame-count strategy, representing each video with a constant number of frames regardless of duration.
During training, GroundVTS-Q uses a frame rate of 2 FPS, while GroundVTS-I samples 16 frames per video.
For the VTS module, the hidden dimension $D_r$ is set to 512 for GroundVTS-Q and 128 for GroundVTS-I.
The visual token sampling ratio is fixed at $\rho\!=\!0.5$.
Additional training settings are detailed in the supplementary material.

\subsection{Main Results}\label{sec:exp_main}
{\bf Moment Retrieval.}
As summarized in Tables~\ref{tab:main_1} and~\ref{tab:main_2}, our proposed GroundVTS consistently outperforms existing state-of-the-art methods on multiple VTG benchmarks.
%demonstrating strong generalization and robustness
\begin{table}[!t]
\centering
\caption{Comparison with state-of-the-art methods on QVHighlights validation split.}
\label{tab:main_2}
\resizebox{\linewidth}{!}{%
\begin{tabular}{l|*{2}{>{$}l<{$}}|*{2}{>{$}l<{$}}}
\toprule
\multirow{2}{*}{Method} & \multicolumn{2}{c|}{MR} & \multicolumn{2}{c}{HD} \\ 
                       & \text{R1@.5} & \text{R1@.7} & \text{mAP} & \text{Hit@1} \\ 
\midrule
\color{gray}{SeViLA$^{\circ}$}\cite{yu2023self}   & \color{gray}{54.5}  & \color{gray}{36.5} & \color{gray}{--}  & \color{gray}{--}\\
\color{gray}{UniVTG$^{\circ}$}\cite{lin2023univtg}                 & \underline{\color{gray}{58.9}}  &  \textbf{\color{gray}{40.9}} & \color{gray}{27.0} & \color{gray}{55.3} \\ 
\midrule
VTG-LLM\cite{guo2025vtg}    & --  & -- & 16.5 & 33.5 \\ 
TimeChat\cite{ren2024timechat}  & --  & -- &  14.5 &  23.9 \\ 
NumPro\cite{wu2025number}  & --  & -- & 40.5 & \underline{70.7} \\ 
\midrule
Qwen2.5VL-7B                 & 8.7  & 2.4 & 24.9 & 0.6 \\ 
Qwen2.5VL-7B-G       & 11.0 & 4.3 & 34.4 & 44.5 \\ 
\rowcolor{gray!15}
GroundVTS-Q        & 23.6_{\color{teal}(\uparrow 12.6)} & 12.3_{\color{teal}(\uparrow 8.0)} & \underline{35.7}_{\color{teal}(\uparrow 1.3)} & 58.8_{\color{teal}(\uparrow 14.3)} \\ 
\midrule
InternVL3.5-8B               & 8.7  & 3.7 & 24.8 & 0.32 \\ 
InternVL3.5-8B-G     & 31.8 & 15.0 & 31.9 & 39.8 \\ 
\rowcolor{gray!15}
GroundVTS-I      & \textbf{63.6}_{\color{teal}(\uparrow 31.8)} & \underline{40.7}_{\color{teal}(\uparrow 25.7)} & \textbf{52.5}_{\color{teal}(\uparrow 20.6)} & \textbf{88.4}_{\color{teal}(\uparrow 48.6)} \\ 
\bottomrule
\end{tabular}%
}
\vspace{3pt}
\footnotesize{%
\begin{minipage}{\linewidth}
\raggedright
$^{\circ}$ indicates classical expert models; other notations follow Table~\ref{tab:main_1}.
\end{minipage}
}
%\vspace{-1em}
\end{table}
%{\color{teal}$\uparrow$} denotes improvement over the corresponding `-G' baseline.
On Charades-STA, GroundVTS-Q substantially outperforms the fine-tuned Qwen2.5VL-7B baseline, achieving gains of 24.8 points in R1@0.5 and 18.4 points in mIoU, reaching 57.5 R1@0.5 and 50.1 mIoU.
On ActivityNet-Captions, GroundVTS-Q improves R1@0.5 by 9.7 points and mIoU by 9.3 points, further confirming the effectiveness of our sampling approach for VTG.
Building upon InternVL3.5-8B, GroundVTS-I also shows stable improvements (\emph{e.g.}, +3.5 in R1@0.7 on Charades-STA), validating the generality of our approach across diverse Vid-LLM architectures.
On QVHighlights, GroundVTS-I attains 63.6 in R1@0.5 and 40.7 in R1@0.7 for moment retrieval, comparable to specialized methods such as UniVTG~\cite{lin2023univtg}.

\noindent{\bf Highlight Detection.}
% As shown in Table~\ref{tab:main_2}, GroundVTS-I achieves substantial improvements over InternVL-G, raising mAP and Hit@1 to 52.5 and 88.4, respectively (+20.6 and +48.6).
% It also surpasses strong baselines such as NumPro~\cite{wu2025number} (40.5 mAP and 70.7 Hit@1),
% demonstrating superior key moment sensitivity and enhanced temporal localization accuracy.
As shown in Table~\ref{tab:main_2}, GroundVTS-I significantly outperforms InternVL3.5-8B-G, improving mAP and Hit@1 by 20.6 and 48.6 points, respectively, to 52.5 and 88.4.
It also surpasses strong methods using frame indices as auxiliary inputs, such as NumPro~\cite{wu2025number}, suggesting better sensitivity to key moments in highlight detection.

%=====%
\begin{table}[!t]
\centering
\caption{Comparison with state-of-the-art methods on NExT-GQA test splits.}
% \caption{State-of-the-art comparison on the NExT-GQA test splits.}
\label{tab:main_3}
\resizebox{\linewidth}{!}{%
\begin{tabular}{@{}l|*{5}{>{$}c<{$}}@{}}
\toprule
Model            & \text{mIoU} & \text{mIoP} & \text{IoU@.5} & \text{IoP@.5} & \text{Acc@GQA} \\ \midrule
\color{gray}{TOGA$^{\circ}$}~\cite{yourcitation2025toga}    & \color{gray}{\underline{24.4}} & \color{gray}{\textbf{40.5}} & \color{gray}{\textbf{21.1}} & \color{gray}{\textbf{40.6}} & \color{gray}{\textbf{24.6}} \\
VideoStreaming~\cite{qian2024streaming}    & 19.3 & 32.2 & 13.3 & 31.0 & 17.8 \\
\midrule
% \rowcolor{gray!15}
GroundVTS-Q        & \textbf{25.8} & \underline{37.4} & \underline{20.4} & \underline{35.4} & \underline{23.2} \\
% \midrule
% \rowcolor{gray!15}
GroundVTS-I     & 16.7 & 26.5 & 11.9 & 24.3 & 18.5 \\ \bottomrule
\end{tabular}%
}
\end{table}
%=====%

\noindent{\bf Out-of-Distribution Evaluation.}
To further assess the effectiveness and generality of our method under task shift, we evaluate our models \emph{as-is} on grounded video question answering with NExT-GQA~\cite{xiao2024can}, without any further training; results are shown in Table~\ref{tab:main_3}.
GroundVTS-Q achieves the highest mIoU and remains competitive on other metrics, despite not being specifically designed or trained for this task.
Moreover, the supplementary material reports two additional \emph{as-is} evaluations: DiDeMo~\cite{anne2017didemo} for out-of-distribution moment retrieval, and LongVideoBench~\cite{wu2024longvideobench} for transfer to a new long-video understanding task; on both benchmarks, our models either outperform or remain competitive with recent state-of-the-art methods.

%------------------------------------------------------
\subsection{Effect of Visual Token Density}\label{sec:top_k}
To verify the robustness of our GroundVTS with respect to visual token density, we conduct an analysis on the Charades-STA test split, comparing GroundVTS-Q with its fine-tuned base model, Qwen2.5VL-7B-G (abbreviated as QwenVL-G).
We adjust the sampling ratio $\rho$ from 0.1 to 1.0 in increments of 0.1, while keeping the dense sampling frame rate fixed at 2 FPS, to control the number of visual tokens involved in LLM inference. For a fair comparison, QwenVL-G continues to use uniform frame sampling, with the frame rate varied from 0.2 to 2 FPS in increments of 0.2.

\begin{figure}
  \centering
  \begin{subfigure}{\linewidth}
    \includegraphics[width=\linewidth]{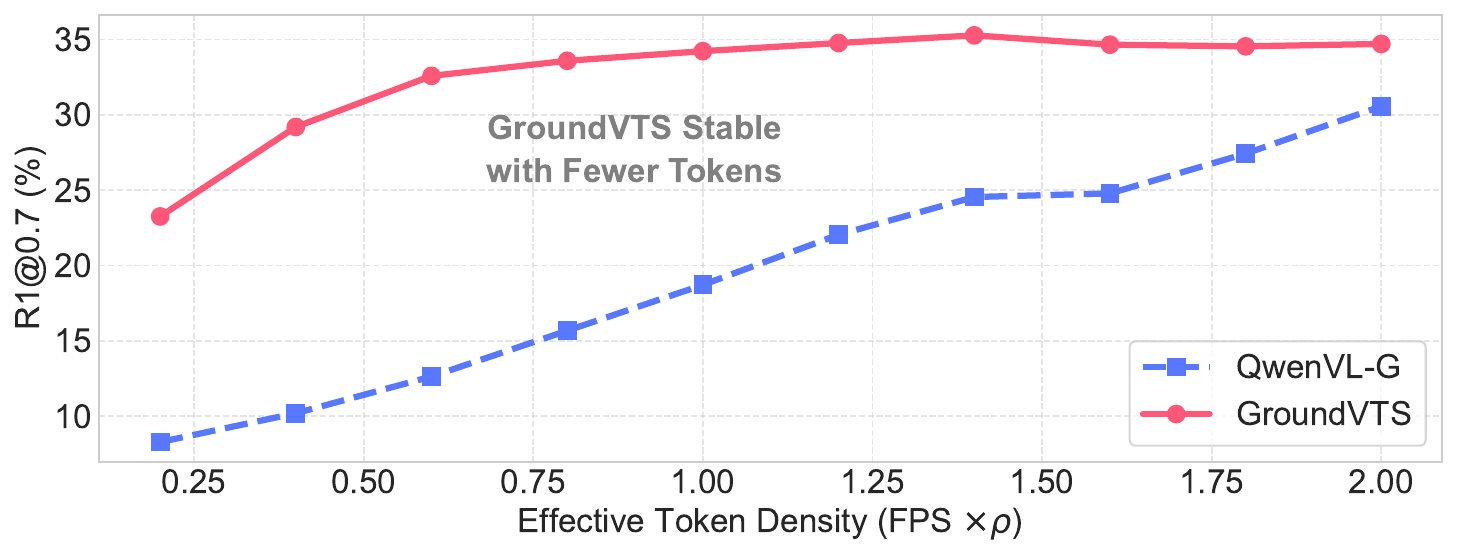}
    \caption{Impact of token density on VTG performance.}
    \label{fig:stable}
  \end{subfigure}
  \begin{subfigure}{\linewidth}
    \includegraphics[width=\linewidth]{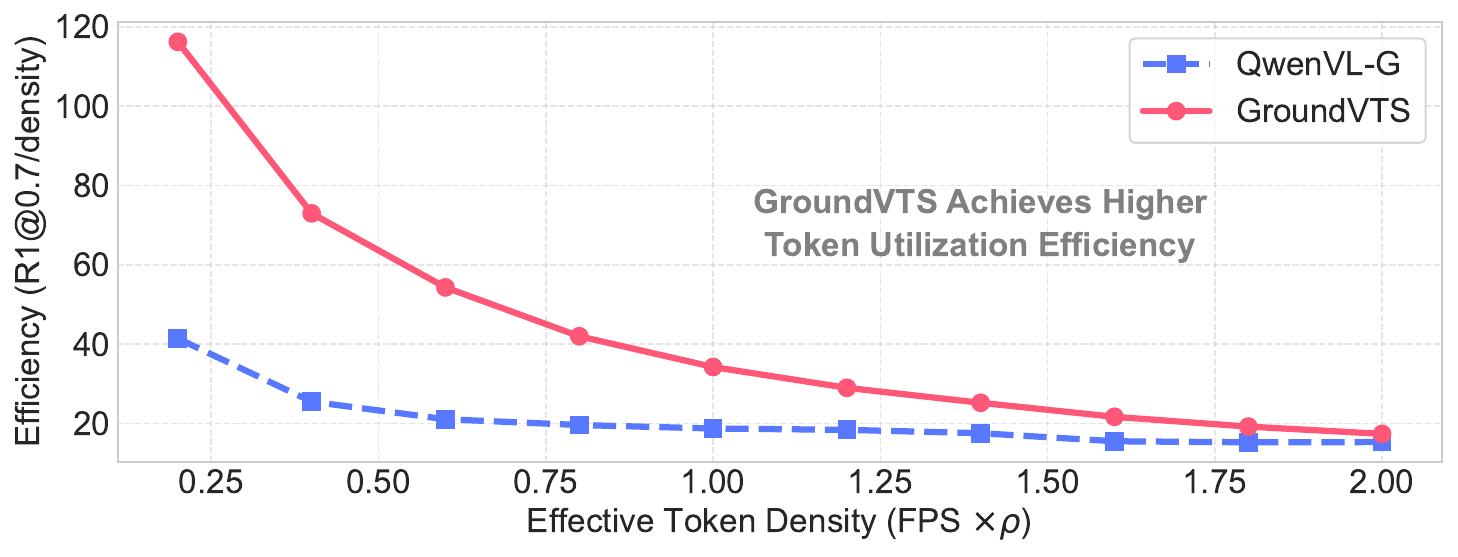}
    \caption{Token efficiency under varying token densities.}
    \label{fig:efficiency}
  \end{subfigure}
  \caption{Comparison between GroundVTS-Q and Qwen2.5VL-7B-G (denoted as QwenVL-G) under varying token densities.}
  \label{fig:density}
\end{figure}

As shown in Figure~\ref{fig:density}\subref{fig:stable}, the horizontal axis represents effective token density ($\text{FPS}\!\times\!\rho$), and the vertical axis reports grounding accuracy in terms of R1@0.7.
As the token density decreases, QwenVL-G degrades markedly, indicating a strong dependence on dense temporal sampling.
In contrast, GroundVTS remains much more stable across the full density range, maintaining high accuracy even in sparse settings.
With only half the token budget ($\text{FPS}\!\times\!\rho\!=\!1.0$), GroundVTS achieves 34.2 R1@0.7, already surpassing QwenVL-G at full density (30.5 R1@0.7).
Even under a more aggressive reduction ($\text{FPS}\!\times\!\rho\!=\!0.4$), GroundVTS still attains 29.2 R1@0.7, exceeding QwenVL-G by 19.0 points.
These results highlight the strong token efficiency and robustness of GroundVTS under sparse sampling.

Figure~\ref{fig:density}\subref{fig:efficiency} illustrates token efficiency, defined as R1@0.7 divided by effective token density.
When fewer visual tokens are available, GroundVTS-Q maintains higher efficiency than QwenVL-G, indicating more effective use of limited visual information.
Exact values are provided in the supplementary material.

%------------------------------------------------------
\subsection{Effect of the Progressive Optimization Strategy}
Table~\ref{tab:train_stage} summarizes the effect of different training stages for GroundVTS-Q.
Using an untrained VTS module (the ``None'' setting) causes a sharp drop across all metrics, showing that query-conditioned token sampling must be properly learned.
Stage 1 (VTS Warm-up) largely recovers the base-model performance, indicating that VTS can be integrated without disrupting the original pipeline.
Adding Stage 2 (Joint LoRA Adaptation) further improves performance, bringing gains of +11.6 in R1@0.3 and +8.0 in mIoU over the base model.
Adding Stage 3 (Grounding Fine-tuning) yields the best results, reaching 71.5/57.5/34.2 R1@0.3/0.5/0.7 and 50.1 mIoU.
The (1, 3) and (2, 3) variants remain below the full setting, confirming the importance of Stage 2 for large-scale adaptation to non-uniform token distributions and Stage 1 for stable initialization.

\begin{table}[!t]
\centering
\caption{Ablation of different training-stage combinations for GroundVTS-Q on Charades-STA test split.}
\label{tab:train_stage}
\resizebox{\linewidth}{!}{%
\begin{tabular}{@{}l|*{4}{>{$}l<{$}}@{}}
\toprule
Stage & \text{R1@0.3} & \text{R1@0.5} & \text{R1@0.7} & \text{mIoU} \\ 
\midrule
base$^{\ddagger}$     & 34.2 & 18.8 & 8.6 & 22.1 \\ 
\midrule
None$^{\mathsection}$       & 8.6_{\color{purple}(\downarrow 25.6)} & 5.0_{\color{purple}(\downarrow 13.8)} & 1.9_{\color{purple}(\downarrow 6.7)} & 5.6_{\color{purple}(\downarrow 16.5)} \\
1       & 31.2_{\color{purple}(\downarrow 3.0)} & 20.5_{\color{teal}(\uparrow 1.7)} & 10.0_{\color{teal}(\uparrow 1.4)} & 20.9_{\color{purple}(\downarrow 1.2)} \\
1, 2    & 45.8_{\color{teal}(\uparrow 11.6)} & 28.8_{\color{teal}(\uparrow 10.0)} & 13.2_{\color{teal}(\uparrow 4.6)} & 30.1_{\color{teal}(\uparrow 8.0)} \\
1, 3     & 49.1_{\color{teal}(\uparrow 14.9)} & 32.5_{\color{teal}(\uparrow 13.7)} & 15.2_{\color{teal}(\uparrow 6.6)} & 32.4_{\color{teal}(\uparrow 10.3)} \\
2, 3     & \underline{69.4}_{\color{teal}(\uparrow 35.2)} & \underline{53.0}_{\color{teal}(\uparrow 34.2)} & \underline{30.5}_{\color{teal}(\uparrow 21.9)} & \underline{47.4}_{\color{teal}(\uparrow 25.3)} \\
1, 2, 3    & \textbf{71.5}_{\color{teal}(\uparrow 37.3)} & \textbf{57.5}_{\color{teal}(\uparrow 38.7)} & \textbf{34.2}_{\color{teal}(\uparrow 25.6)} & \textbf{50.1}_{\color{teal}(\uparrow 28.0)} \\
\bottomrule
\end{tabular}
}
\vspace{3pt}
\footnotesize{
\begin{minipage}{\linewidth}
\raggedright
$^{\ddagger}$ Arrowed values indicate absolute changes relative to the base model (Qwen2.5VL-7B). $^{\mathsection}$ ``None'' uses a randomly initialized VTS module.
\end{minipage}
}
\end{table}

%------------------------------------------------------
\begin{table*}[!t]
\centering
\small
\caption{Ablation on sampling strategies and positional encoding (PE) in GroundVTS.}
\label{tab:sampling strategy}
% \resizebox{\textwidth}{!}{%
\begin{tabular}{@{}cc|c|cccc|cccc@{}}
\toprule
\multirow{2}{*}{VTS}   & \multirow{2}{*}{PE}  & \multirow{2}{*}{Sampling Methods} & \multicolumn{4}{c|}{Charades-STA}                                                                                & \multicolumn{4}{c}{ActivityNet-Captions}                                                                        \\
&                &                    & \multicolumn{1}{l}{R1@0.3} & \multicolumn{1}{l}{R1@0.5} & \multicolumn{1}{l}{R1@0.7} & \multicolumn{1}{l|}{mIoU} & \multicolumn{1}{l}{R1@0.3} & \multicolumn{1}{l}{R1@0.5} & \multicolumn{1}{l}{R1@0.7} & \multicolumn{1}{l}{mIoU} \\ \midrule
\checkmark & \checkmark & Token-Level                       & \textbf{71.5}                     & \textbf{57.5}                      & \textbf{34.2}                       & \textbf{50.1}                      & \textbf{51.3}                       & \textbf{33.6}                      & \textbf{21.4}                       & \textbf{36.0}                     \\
\checkmark & \checkmark & Frame-Level                       & \underline{61.7}                       & \underline{44.9}                       & \underline{23.3}                       & \underline{41.6}                      &  \underline{43.7}		
                          &        \underline{27.5}                    &                 \underline{15.0}           &      \underline{30.7}                    \\
-- & \checkmark             & Uniform                           & 42.6                       & 28.5                       & 15.0                       & 29.3                      & 36.1                       & 19.5                       & 7.5                        & 23.4                     \\
-- & \checkmark           & Random                            & 54.9                       & 35.0                       & 16.3                       & 35.7                      & 40.3                       & 23.4                       & 12.1                       & 27.7                     \\
\checkmark & --  & Token-Level                       & 15.1                       & 7.0                       & 2.7                       & 9.5                     & 22.2                       & 11.2                       & 5.2                       & 16.3                     \\ \bottomrule
\end{tabular}%
% }
\end{table*}

%-------------------------------------------------------------------------
\begin{figure*}[!t]
  \centering
  \includegraphics[width=\linewidth]{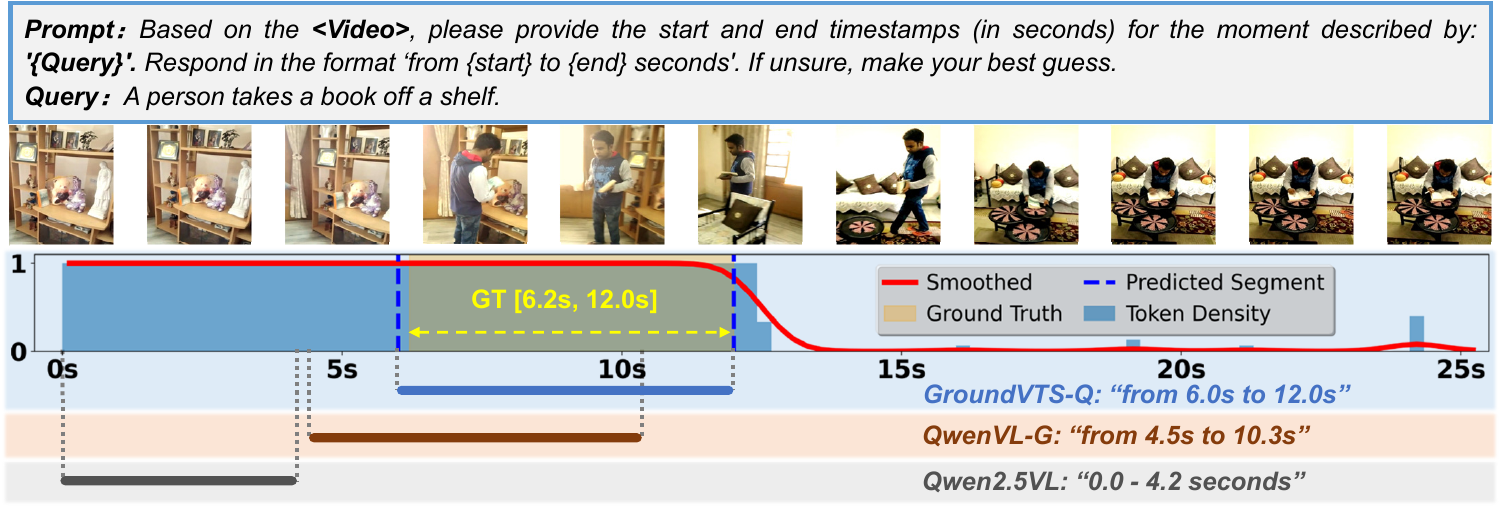}
  \caption{Qualitative comparison of temporal grounding predictions among GroundVTS-Q, Qwen2.5VL-7B-G, and Qwen2.5VL-7B.}
  \label{fig:qualitative}
\end{figure*}
\vspace{-0.1em}
%-------------------------------------------------------------------------
\subsection{Ablation Study}\label{sec:sampling}
We conduct ablation experiments on two key components of GroundVTS in Table~\ref{tab:sampling strategy}: the visual token sampling strategy and the positional encoding used for temporal reasoning.
%All settings follow the same three-stage training procedure, except for the uniform baseline, and are evaluated under a fixed test configuration with $\text{FPS}=2.0$ and sampling ratio $\rho=0.5$.
All variants are evaluated under a matched token budget, equivalent to $\text{FPS}=2.0$ and $\rho=0.5$.

\textbf{Sampling Strategy.}
We compare our query-guided token-level sampling with three alternatives: (a) \textit{Uniform sampling}, implemented by evaluating Qwen2.5VL-7B-G at 1.0 FPS; (b) \textit{Random sampling}, where 50\% of visual tokens are randomly discarded; and (c) \textit{Frame-level query selection}, where visual tokens within each frame are average-pooled to estimate frame-query relevance, and the top 50\% frames are retained with all their tokens.
Both the token-level and frame-level variants are trained with the same three-stage procedure, while the random variant is initialized from the token-level model after Stages 1 and 2 and trained only in Stage 3 with random dropping.

As shown in Table~\ref{tab:sampling strategy}, our token-level VTS achieves the best performance on both datasets.
On Charades-STA, it improves mIoU over frame-level selection by 8.5 points (50.1 vs.\ 41.6) and reaches 57.5 R1@0.5.
On ActivityNet-Captions, it also performs best, improving mIoU by 5.3 points over the frame-level variant (36.0 vs.\ 30.7).
By contrast, both uniform and random sampling degrade performance, confirming the importance of query-guided fine-grained sampling for temporal grounding; random sampling nevertheless outperforms uniform sampling, possibly because it acts as data augmentation.
%-----------
%We hypothesize that random sampling acts as a form of data augmentation, contributing to its gains over uniform sampling.
%-----------

\textbf{Effect of Positional Encoding.}
To assess the role of positional encoding, we remove position embeddings from GroundVTS while keeping training and inference settings fixed.
As shown in Table~\ref{tab:sampling strategy}, performance collapses on both datasets.
On Charades-STA, mIoU drops from 50.1 to 9.5 and R1@0.5 from 57.5 to 7.0, with similarly severe degradation on ActivityNet-Captions.
These results confirm the importance of temporal positional information in GroundVTS, and validate our design choice of retaining the original relative positional embeddings for the selected tokens.

\textbf{Additional ablations} on training data and relevance estimation are provided in the supplementary material.
%-------------------------------------------------------------------------

\subsection{Qualitative Study}\label{sec:qualitative}
Figure~\ref{fig:qualitative} shows a qualitative comparison on a Charades-STA example with the query ``a person takes a book off a shelf.''
The red curve denotes the normalized token density produced by VTS, with higher values indicating stronger query relevance.
GroundVTS-Q assigns most tokens to the early part of the video (roughly 0--13 s), which fully covers the ground-truth interval (6.2--12.0 s), while suppressing nearly all tokens in later frames.
Based on these sampled tokens, GroundVTS-Q predicts 6.0--12.0 s, closely matching the ground truth.
In contrast, Qwen2.5VL-7B-G predicts an earlier and less precise segment (4.5--10.3 s), while the base Qwen2.5VL-7B misses the target moment entirely.
%This example shows that VTS effectively filters out irrelevant content while preserving the informative temporal region needed for accurate grounding.
This shows that VTS focuses on relevant temporal regions for grounding.
More results are provided in the supplement.
\section{Conclusion}
\label{sec:conclusion}

In this paper, we present GroundVTS, a query-guided visual token sampling framework for video temporal grounding.
Its core module, VTS, can be seamlessly integrated into mainstream Vid-LLMs via a progressive optimization strategy to better capture fine-grained temporal cues.
Experiments show that GroundVTS consistently improves instruction-tuned base models and outperforms recent state-of-the-art methods.
Further analyses confirm that GroundVTS improves token utilization and maintains prediction stability across varying input densities.
%Overall, GroundVTS offers a practical and generalizable way to improve fine-grained temporal reasoning in Vid-LLMs.

\vspace{1ex}\noindent\textbf{Acknowledgements}.
% This paper is in part supported by the National Natural Science Foundation of China under Grant 624B2105.
The research of Liuyi Wang is supported in part by the National Natural Science Foundation of China under Grant 624B2105.
%\newpage
%\input{sec/6_ack}

% WARNING: do not forget to delete the supplementary pages from your submission 
% \input{sec/X_suppl}

{
    \small
    \bibliographystyle{ieeenat_fullname}
    \bibliography{main}
}

% \appendix
\clearpage
\setcounter{page}{1}
\maketitlesupplementary

\section{Influence on General VQA}\label{sec:GVQA}

\begin{table*}[!t]
\centering
\small
\caption{Comparison with base model on MVBench subtasks.}
\label{tab:GVQA}
\begin{tabular}{@{}l|ccccccccccc|c@{}}
\toprule
Model       & AA   & AC & AL & AS   & EN   & ER   & FGA  & OI   & OS   & ST   & SC   & all  \\ \midrule
Qwen2.5-VL-7B & 77   & 39 & 38 & 69.7 & 30   & 50   & 44.5 & 66.5 & 35.5 & 90.5 & 50.5 & 53.7 \\ 
GroundVTS-Q   & 75.5 & 47 & 46 & 70.2 & 28.5 & 46.5 & 41   & 63.5 & 43   & 89.5 & 48.5 & 54.5 \\ \bottomrule
\end{tabular}
\end{table*}

While GroundVTS significantly improves temporal grounding accuracy, it is essential to evaluate whether its selective token sampling mechanism impacts the model's ability to handle general video understanding tasks. To investigate this, we assess GroundVTS on multiple subtasks of MVBench~\cite{li2024mvbench}, a benchmark that comprehensively measures various video question-answering (VQA) capabilities, including temporal reasoning, object interaction, and scene-level comprehension.

As shown in Table~\ref{tab:GVQA}, GroundVTS-Q achieves a slightly higher overall score (+0.8) compared to the base model Qwen2.5-VL-7B, indicating that the introduction of the VTS module does not compromise general VQA capability. 
The overall performance of GroundVTS-Q is competitive, with significant improvements in several tasks.
The most notable gains appear in subtasks that focus on temporal reasoning and fine-grained action reasoning, such as Action Count (AC), Action Localization (AL) and Action Sequence (AS). In these tasks, GroundVTS-Q outperforms the base model by a substantial margin, with an improvement of 8.0 points in AC, 8.0 points in AL and 0.5 points in AS. These results align with the design objective of enhancing temporal sensitivity, confirming that VTS excels at capturing the temporal aspects of video data.

Meanwhile, in tasks that require a more general understanding of the scene or object interactions, such as Scene Transition (ST) and State Change (SC), GroundVTS-Q maintains competitive performance. For example, GroundVTS-Q achieves a slight drop in ST (-1.0) and SC (-2.0) compared to the base model, but still performs well overall. This suggests that the selective filtering mechanism of the VTS module does not undermine the model's ability to grasp global scene awareness or appearance-related information.

In summary, the results suggest that the introduction of VTS enhances the model’s performance on tasks requiring fine-grained temporal reasoning, while maintaining strong performance in more general video understanding tasks. This balance between focused temporal sensitivity and general visual understanding demonstrates the versatility and effectiveness of the GroundVTS framework.

\section{Out-of-distribution Data Experiment}

To further evaluate the robustness and generalization ability of GroundVTS, we conduct out-of-distribution (OOD) experiments on three benchmarks: DiDeMo~\cite{anne2017didemo}, LongVideoBench~\cite{wu2024longvideobench}, and NExT-GQA~\cite{xiao2024can}. None of these datasets are included in the fine-tuning data of either GroundVTS or its base models (Qwen2.5-VL and InternVL3.5). This evaluation examines whether our method can effectively transfer its temporal grounding capability to unseen domains and tasks.

\textbf{Results on DiDeMo.}
As shown in Table~\ref{tab:ood1}, GroundVTS-Q achieves substantial gains over both the instruction-tuned baseline QwenVL-G and the pretrained Qwen2.5-VL. Specifically, it improves by +13.7 R1@0.3, +10.1 R1@0.5, and +8.0 mIoU, establishing new SOTA performance on two of the three metrics. These results highlight the strong cross-domain adaptability brought by our query-guided VTS mechanism.
On InternVL3.5-8B, GroundVTS-I also produces consistent improvements over InternVL-G, yielding gains of +0.9 R1@0.3, +1.3 R1@0.5, and +1.1 mIoU. While the improvements are smaller than those observed with Qwen2.5-VL, they confirm that GroundVTS remains effective even when the underlying base model already possesses strong temporal reasoning capability.

\textbf{Results on LongVideoBench.}
LongVideoBench evaluates temporal reasoning on significantly longer videos, with durations ranging from tens of seconds to several minutes. As shown in Table~\ref{tab:ood2}, GroundVTS-I achieves the best accuracy on two of the three duration ranges, reaching 65.6\% on (8,15]s and 68.0\% on (15,60]s, and remains competitive on (180,600]s. These results suggest that the proposed query-guided temporal selection mechanism can effectively scale to long-video scenarios by filtering irrelevant temporal regions and concentrating computation on query-related segments. 

\textbf{Results on NExT-GQA.}
NExT-GQA evaluates both temporal grounding and question answering accuracy, requiring models to first localize relevant temporal segments before answering questions. As shown in Table~\ref{tab:ood3}, GroundVTS-Q achieves the best mIoU (25.8) among all compared methods and remains competitive on other grounding-related metrics. Notably, the strongest baseline, TOGA, is a classical expert model specifically designed for grounded video question answering, whereas GroundVTS is built upon a general-purpose multimodal large language model without task-specific architecture. Despite this difference, GroundVTS-Q still attains comparable performance across most metrics, demonstrating that the proposed query-guided temporal selection mechanism can effectively transfer its temporal localization capability to reasoning-intensive video QA tasks.

Overall, the OOD evaluation across three diverse benchmarks demonstrates that GroundVTS generalizes well to unseen datasets and tasks, including short video grounding (DiDeMo), long-video reasoning (LongVideoBench), and grounded video question answering (NExT-GQA). These results reinforce the robustness of the proposed framework and its ability to capture transferable fine-grained temporal cues beyond the training distribution.

\begin{table}[!t]
\centering
\caption{Comparison with state-of-the-art methods on DiDeMo test splits.}
\label{tab:ood1}
\resizebox{\linewidth}{!}{
\begin{tabular}{@{}l|*{3}{>{$}l<{$}}@{}}
\toprule
Model            & \text{R1@0.3} & \text{R1@0.5} & \text{mIoU}  \\ \midrule
Video-LLaMA~\cite{zhang2023video}    & 20.1 & 8.2 & 14.3 \\
Video-ChatGPT~\cite{maaz2024video}        & 19.8 & 6.5 & 13.7 \\
Valley     & 33.2 & 13.4 & 21.8     \\ 
VideoChat~\cite{li2023videochat}   & 34.5 & 14.5 & 22.4  \\ 
Momenter    & 38.2 & 21.8 & 26.5  \\ 
VTimeLLM~\cite{huang2024vtimellm}    & \underline{45.0} & \underline{28.8} & 27.9  \\ 
TimeChat~\cite{ren2024timechat}     & 42.8 & 24.4 & 28.2 \\ 
HawkEye      & 44.8 & \textbf{29.7} & \underline{29.5} \\ 
\midrule
Qwen2.5-VL-7B    & 28.7	& 22.7	& 22.2 \\
QwenVL-G        & 32.6	& 17.7	& 22.0 \\
GroundVTS-Q     & \textbf{46.3}_{\color{teal}(\uparrow 13.7)}	& 27.8_{\color{teal}(\uparrow 10.1)}   & \textbf{30.0}_{\color{teal}(\uparrow 8.0)}    \\ \midrule
InternVL3.5-8B       & 29.5	& 23.9	& 23.0 \\
InternVL-G      & 36.6	& 21.0	& 23.1\\
GroundVTS-I     & 37.5_{\color{teal}(\uparrow 0.9)}	& 22.3_{\color{teal}(\uparrow 1.3)}	& 24.2_{\color{teal}(\uparrow 1.1)} \\ \bottomrule
\end{tabular}%
}
\vspace{3pt}
\footnotesize
\begin{minipage}{\linewidth}
\raggedright
The baseline results are from reference~\cite{fang2025and}.
\end{minipage}
\end{table}

% \begin{table}[!t]
% \centering
% \caption{Comparison with state-of-the-art methods on LongVideo-Bench test splits (Acc).}
% \label{tab:ood2}
% \resizebox{\linewidth}{!}{%
% \begin{tabular}{@{}l|lll@{}}
% \toprule
% Model            & (8, 15]s & (15, 60]s & (180, 600]s  \\ \midrule
% VideoTree~\cite{wang2025videotree}    & 61.0   & 57.5   & 48.4 \\
% VideoMiner~\cite{cao2025videominer}    & \underline{65.1}   & \underline{64.7}   & \textbf{58.6} \\
% \midrule
% GroundVTS-Q        & 52.9 & 60.5 & 44.2 \\
% GroundVTS-I     & \textbf{65.6} & \textbf{68.0} & \underline{52.4} \\ \bottomrule
% \end{tabular}%
% }
% \end{table}

\begin{table}[!t]
\centering
\small
\caption{Comparison with state-of-the-art methods on LongVideo-Bench test splits (Acc).}
\label{tab:ood2}
\begin{tabular}{@{}l|ccc@{}}
\toprule
Model & (8, 15]s & (15, 60]s & (180, 600]s \\ 
\midrule
VideoTree~\cite{wang2025videotree} & 61.0 & 57.5 & 48.4 \\
VideoMiner~\cite{cao2025videominer} & \underline{65.1} & \underline{64.7} & \textbf{58.6} \\
\midrule
GroundVTS-Q & 52.9 & 60.5 & 44.2 \\
GroundVTS-I & \textbf{65.6} & \textbf{68.0} & \underline{52.4} \\
\bottomrule
\end{tabular}
\end{table}

\begin{table}[!t]
\centering
\caption{Comparison with state-of-the-art methods on NExT-GQA test splits.}
\label{tab:ood3}
\resizebox{\linewidth}{!}{%
\begin{tabular}{@{}l|ccccc@{}}
\toprule
Model            & mIoU & mIoP & IoU@.5 & IoP@.5 & Acc@GQA \\ \midrule
\color{gray}{TOGA}~\cite{yourcitation2025toga}    & \color{gray}{\underline{24.4}} & \color{gray}{\textbf{40.5}} & \color{gray}{\textbf{21.1}} & \color{gray}{\textbf{40.6}} & \color{gray}{\textbf{24.6}} \\
VidStreaming~\cite{qian2024streaming}    & 19.3 & 32.2 & 13.3 & 31.0 & 17.8 \\
\midrule
GroundVTS-Q        & \textbf{25.8} & \underline{37.4} & \underline{20.4} & \underline{35.4} & \underline{23.2} \\
GroundVTS-I     & 16.7 & 26.5 & 11.9 & 24.3 & 18.5 \\ \bottomrule
\end{tabular}%
}
\end{table}

\section{Parameter-free projection}\label{sec:parameter_free}
We evaluate a parameter-free relevance estimation method (Table~\ref{tab:parameter_free}). Without additional training, this approach leads to a substantial performance drop due to the mismatch between the sampled-token distribution and the pretrained LLM (w/o training). To alleviate this issue, we further fine-tune the LLM following Stages 2\&3 to adapt to the sampled-token distribution (w/ training), which partially recovers the performance. However, it still underperforms the full GroundVTS model with learned relevance projections.

\begin{table*}[!t]
\centering
\small
\caption{Parameter-free token sampling vs. GroundVTS.}
\label{tab:parameter_free}
% \resizebox{\textwidth}{!}{%
\begin{tabular}{@{}cc|cccc|cccc@{}}
\toprule
\multirow{2}{*}{VTS}  & \multirow{2}{*}{Training}  & \multicolumn{4}{c|}{Charades-STA}                                                                                & \multicolumn{4}{c}{ActivityNet-Captions}                                                                        \\
&                & \multicolumn{1}{l}{R1@0.3} & \multicolumn{1}{l}{R1@0.5} & \multicolumn{1}{l}{R1@0.7} & \multicolumn{1}{l|}{mIoU} & \multicolumn{1}{l}{R1@0.3} & \multicolumn{1}{l}{R1@0.5} & \multicolumn{1}{l}{R1@0.7} & \multicolumn{1}{l}{mIoU} \\ \midrule
\checkmark & \checkmark  & \textbf{71.5}                     & \textbf{57.5}                      & \textbf{34.2}                       & \textbf{50.1}                      & \textbf{51.3}                       & \textbf{33.6}                      & \textbf{21.4}                       & \textbf{36.0}                     \\
-- & \checkmark          & 69.6 & 52.7 & 29.6 & 47.5 & 38.8 & 25.0 & 13.7 & 27.8                    \\
--  & --         & 21.2 & 13.6 & 6.8 & 14.5 & 9.1 &  5.1 & 2.6 & 6.6                    \\\bottomrule
\end{tabular}%
% }
\end{table*}

\section{Dataset Ablation}

GroundVTS-Q is trained on the LLaVA-Video-178K and our constructed Grounding-FT datasets. 
For a fair comparison, the base Qwen2.5VL-7B is trained on the same datasets under three settings, as reported in Table~\ref{tab:dataset_ablation}: 
(i) \textit{Qwen-G}, trained only on the Grounding-FT dataset; 
(ii) \textit{Qwen-(L+G)}, trained on the concatenation of the two datasets; and 
(iii) \textit{Qwen-(L$\shortrightarrow$G)}, trained following the same Stage 2$\shortrightarrow$Stage 3 curriculum. 
Note that Stage 1 (VTS warm-up) is not applicable to the base model.

As shown in Table~\ref{tab:dataset_ablation}, GroundVTS-Q consistently outperforms the Qwen baselines across all evaluation settings. In particular, GroundVTS-Q achieves 50.1 mIoU on Charades-STA, significantly surpassing Qwen-G with 31.7, Qwen-(L+G) with 28.5, and Qwen-(L$\shortrightarrow$G) with 29.8. These results suggest that the proposed grounding-aware training strategy effectively improves performance under matched data and training configurations.

\begin{table}[!t]
\centering
\footnotesize
\setlength{\tabcolsep}{4pt}
\caption{Dataset ablation on Charades-STA and ActivityNet-Captions test spilt.}
\label{tab:dataset_ablation}

\begin{tabularx}{\linewidth}{@{}lcccc cccc@{}}
\toprule
\multirow{2}{*}{Variant} & \multicolumn{4}{c}{Charades-STA} & \multicolumn{4}{c}{ActivityNet-Captions} \\
\cmidrule(lr){2-5} \cmidrule(lr){6-9}
 & {\footnotesize R1} & {\footnotesize R1} & {\footnotesize R1} & \multirow{2}{*}{\footnotesize mIoU} 
 & {\footnotesize R1} & {\footnotesize R1} & {\footnotesize R1} & \multirow{2}{*}{\footnotesize mIoU} \\
 & {\footnotesize @.3} & {\footnotesize @.5} & {\footnotesize @.7} 
 & 
 & {\footnotesize @.3} & {\footnotesize @.5} & {\footnotesize @.7} 
 & \\
\midrule
Qwen-G                      & 45.2 & 32.7 & 18.7 & 31.7 & 40.6 & 23.9 & 9.9 & 26.7 \\
Qwen-(L{$+$}G)                & 41.1 & 27.7 & 15.7 & 28.5 & 39.1 & 20.6 & 7.8 & 24.9 \\
Qwen-(L{$\shortrightarrow$}G) & 42.5 & 30.7 & 16.9 & 29.8 & 40.0 & 22.1 & 8.6 & 25.9 \\
GroundVTS-Q & \textbf{71.5}  & \textbf{57.5}  & \textbf{34.2}  & \textbf{50.1}    & \textbf{51.3}  & \textbf{33.6}  & \textbf{21.4} & \textbf{36.0} \\
\bottomrule
\end{tabularx}
\end{table}

\section{Frame Sampling Sensitivity of InternVL3.5}\label{sec:intern_sensitivity}

To examine whether the frame density sensitivity is specific to Qwen2.5-VL or reflects a more general phenomenon, we conduct an additional experiment using InternVL3.5. 
Unlike Qwen2.5-VL, InternVL3.5 adopts a fixed-number frame sampling strategy. Therefore, we vary the number of sampled frames to analyze how visual token density affects VTG performance.

\begin{figure}[!t]
  \centering
  \includegraphics[width=\linewidth]{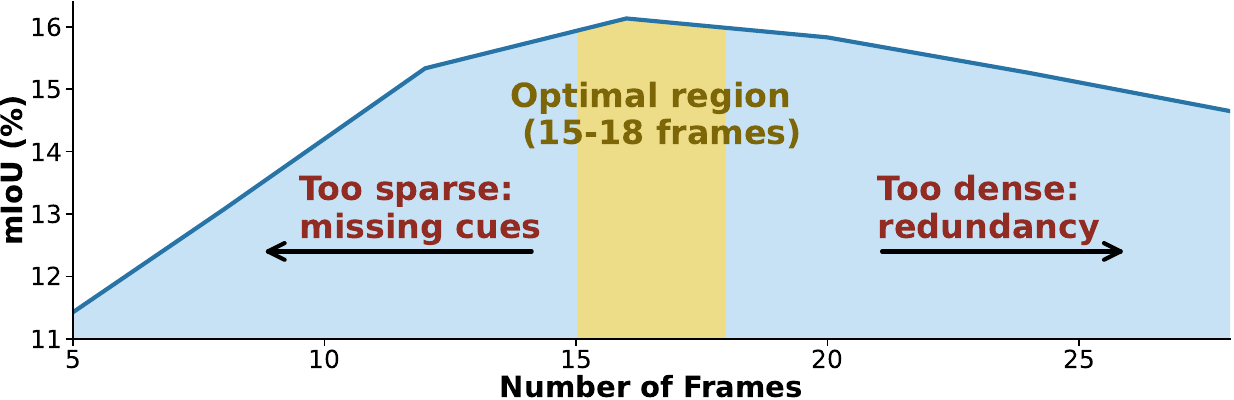}
  \caption{Frame sensitivity of InternVL3.5 on QVHighlights.}
  \label{fig:fn_sensitivity}
\end{figure}

Figure~\ref{fig:fn_sensitivity} presents the frame sensitivity results of InternVL3.5 on the QVHighlights dataset.
Similar to the trend observed with Qwen2.5-VL, the performance again exhibits a clear non-linear dependency on frame density. Increasing the number of sampled frames initially improves performance by providing richer temporal cues. However, beyond a certain point, further increasing the frame count leads to diminishing returns and eventually performance degradation, suggesting that excessive visual tokens introduce redundancy and interfere with effective temporal reasoning.

These results indicate that the sensitivity to visual token density is not limited to a specific model architecture, but appears to be a general characteristic of multimodal LLM-based VTG systems. This observation further supports our motivation for designing an adaptive token sampling mechanism in GroundVTS.

\section{Additional Analysis of Visual Token Density}\label{sec:token_density_appendix}

Table~\ref{tab:token_density_all} provides the full quantitative results corresponding to the visual token density analysis discussed in the main paper. The results further substantiate the trends previously observed. For the pretrained Qwen2.5VL-7B, performance grows steadily as token density increases, but drops rapidly in sparse conditions. This confirms its heavy dependence on dense temporal evidence: when the effective density falls below 1.0, all metrics decrease sharply (e.g., R1@0.5 drops from 47.1 to 18.8 as density reduces from 2.0 to 1.0). The fine-tuned QwenVL-G shows improved overall accuracy but remains highly sensitive to token density.

In contrast, GroundVTS-Q demonstrates remarkable stability across all density levels. 
At extremely sparse levels (e.g., density 0.2--0.6), its performance remains comparable to the best performance of its base model, avoiding the sharp degradation observed in QwenVL-G. 
As the density increases, the improvement of GroundVTS-Q is much more gradual, forming a plateau rather than a steep curve. This consistency appears across all evaluation metrics, including R1@0.3, R1@0.5, R1@0.7, and mIoU, illustrating that GroundVTS effectively mitigates the vulnerability of Vid-LLMs to insufficient visual tokens. These results further reinforces the conclusion that our query-guided sampling mechanism yields reliable grounding accuracy regardless of input density, while base Vid-LLMs suffer substantial degradation when token budgets are reduced.

\begin{table*}[!t]
\centering
\caption{Quantitative analysis of visual token density on Charades-STA test split.}
\label{tab:token_density_all}
\resizebox{\textwidth}{!}{%
\begin{tabular}{@{}c|cccc|cccc|cccc@{}}
\toprule
\multirow{2}{*}{$\text{FPS}*\rho$} & \multicolumn{4}{c|}{Qwen2.5VL-7B} & \multicolumn{4}{c|}{QwenVL-G} & \multicolumn{4}{c}{GroundVTS-Q} \\
& R1@0.3 & R1@0.5 & R1@0.7 & mIoU & R1@0.3 & R1@0.5 & R1@0.7 & mIoU & R1@0.3 & R1@0.5 & R1@0.7 & mIoU \\
\midrule
0.2 & 22.7 & 13.3 & 6.8 & 16.1 & 28.0 & 16.5 & 8.3 & 18.7 & 61.2 & 43.6 & 23.3 & 41.0 \\
0.4 & 23.5 & 13.5 & 6.6 & 16.4 & 33.0 & 20.2 & 10.2 & 21.8 & 67.1 & 50.8 & 29.2 & 46.0 \\
0.6 & 26.5 & 14.1 & 6.8 & 17.6 & 36.2 & 24.0 & 12.6 & 24.9 & 69.5 & 54.4 & 32.6 & 48.2 \\
0.8 & 30.3 & 16.4 & 7.2 & 19.7 & 41.4 & 28.3 & 15.7 & 28.4 & 70.9 & 56.8 & 33.6 & 49.6 \\
1.0 & 34.2 & 18.8 & 8.6 & 22.1 & 45.2 & 32.7 & 18.7 & 31.7 & 71.5 & 57.5 & 34.2 & 50.1 \\
1.2 & 36.8 & 21.4 & 9.1 & 24.2 & 49.3 & 36.6 & 22.0 & 35.2 & 72.3 & 58.4 & 34.8 & 50.7 \\
1.4 & 42.2 & 25.0 & 10.6 & 28.0 & 54.8 & 40.8 & 24.5 & 38.9 & 72.9 & 58.5 & 35.3 & 51.0 \\
1.6 & 48.6 & 29.1 & 12.5 & 32.2 & 62.7 & 44.6 & 24.8 & 42.8 & 72.8 & 58.3 & 34.7 & 50.8 \\
1.8 & 59.5 & 37.0 & 16.6 & 38.7 & 72.2 & 53.0 & 27.4 & 48.2 & 73.0 & 58.3 & 34.5 & 50.9 \\
2.0 & 68.8 & 47.1 & 23.5 & 45.4 & 74.4 & 56.3 & 30.6 & 50.0 & 72.8 & 58.5 & 34.7 & 50.9 \\
\bottomrule
\end{tabular}
}
\end{table*}

\section{Training Details}\label{sec:train_config}
This section provides the detailed configurations and parameters used for training GroundVTS across its different stages, as well as the parameter values for the model variants. Table~\ref{tab:training_config} outlines the settings for each stage of training, including the learning rate, optimizer, batch size, and other critical training details. Table~\ref{tab:parameter_statistics} lists the total and trainable parameters for both GroundVTS-Q (Qwen2.5VL-based) and GroundVTS-I (InternVL-based) models.

\begin{table*}[ht]
\centering
\small
\caption{Training configuration for each stage of the GroundVTS model.}
\label{tab:training_config}
\resizebox{\textwidth}{!}{%
\begin{tabular}{@{}c|cccccccc@{}}
\toprule
Stage &
  \begin{tabular}[c]{@{}c@{}}Trainable\\ Modules\end{tabular} &
  \begin{tabular}[c]{@{}c@{}}Learning\\ Rate\end{tabular} &
  Optimizer &
  \begin{tabular}[c]{@{}c@{}}Batch Size\\ (per GPU)\end{tabular} &
  \begin{tabular}[c]{@{}c@{}}Grad. Acc. \\ Steps\end{tabular} &
  Epochs &
  \begin{tabular}[c]{@{}c@{}}LoRA\\ Config\end{tabular} &
  Dataset \\ \midrule
\begin{tabular}[c]{@{}c@{}}Stage 1: VTS\\ Warm-up\end{tabular} &
  VTS &
  1e-5 &
  \multirow{3}{*}[-5ex]{\makecell[c]{AdamW, \\ $\beta_1$ = 0.9, \\ $\beta_2$ = 0.999}} &
  2 &
  4 &
  1 &
  – &
  LLaVA-Video-178K \\
\begin{tabular}[c]{@{}c@{}}Stage 2: Joint\\ LoRA Adaptation\end{tabular} &
  \begin{tabular}[c]{@{}c@{}}LLM (LoRA) +\\  VTS + Projector\end{tabular} &
  2e-4 &
   &
  2 &
  4 &
  2 &
  \begin{tabular}[c]{@{}c@{}}rank = 8,\\ $\alpha$ = 16,\\ dropout = 0.05\end{tabular} &
  LLaVA-Video-178K \\
\begin{tabular}[c]{@{}c@{}}Stage 3: Grounding\\ Fine-tuning\end{tabular} &
  \begin{tabular}[c]{@{}c@{}}LLM (LoRA) + \\ VTS + Projector\end{tabular} &
  1e-4 &
   &
  2 &
  4 &
  3 &
  \begin{tabular}[c]{@{}c@{}}rank = 8,\\ $\alpha$ = 16,\\ dropout = 0.05\end{tabular} &
  Grounding-FT \\ \bottomrule
\end{tabular}%
}
\end{table*}

\begin{table}[ht]
\centering
\small
\caption{Parameter statistics for GroundVTS-Q and GroundVTS-I models.}
\label{tab:parameter_statistics}
\resizebox{\linewidth}{!}{%
\begin{tabular}{@{}l|c|cccc@{}}
\toprule
\multirow{2}{*}{Model} & \multirow{2}{*}{Total  Params} & \multicolumn{4}{c}{Trainable Params} \\
                     &          & VTS & Projector & LoRA & All \\ \midrule
GroundVTS-Q & 8.32B & 29.4M     & 44.6M           & 79.0M      & 153.0M     \\
GroundVTS-I & 8.56B & 34.6M     & 33.6M           & 77.0M      & 145.2M     \\ \bottomrule
\end{tabular}%
}
\end{table}

\section{Grounding-FT Dataset}\label{sec:instruction}

Grounding-FT is a curated dataset designed for instruction fine-tuning on Video Temporal Grounding (VTG) tasks. It aggregates the training splits of Charades-STA~\cite{gao2017tall}, QVHighlights~\cite{lei2021detecting}, and ActivityNet-Captions~\cite{krishna2017dense}, resulting in 70K annotated clips paired with instruction-style queries. The goal is to unify multiple VTG formulations under a consistent question-answering (QA) framework, facilitating language model training with natural conversational inputs rather than fixed task templates.

\subsection{Overview and Construction}
Grounding-FT covers two main VTG task types:

{\bf (a) Moment Retrieval (MR)}---identifying the temporal segment in a video that corresponds to a given natural language query.

{\bf (b) Highlight Detection (HD)}---output all salient moments relevant to the query in the video together with their corresponding relevance scores. 

For MR, we aggregate annotations from the training splits of Charades-STA, QVHighlights, and ActivityNet-Captions. For HD, we use the training split of QVHighlights. All samples are reformulated into an instruction-response style and stored in the {\tt ShareGPT} format, where each instance contains a conversational pair between a {\tt user} (prompt) and an {\tt assistant} (answer), along with the corresponding video path.
To enhance linguistic diversity and improve generalization to natural language instructions, we construct a pool of prompt templates and randomly select one for each instance rather than relying on a single fixed phrasing. This variation helps the fine-tuned model better adapt to free-form human queries.
Note that timestamp information is not provided in the text prompt, and all models must rely on the positional encodings of visual tokens to infer temporal information.

\subsection{Moment Retrieval Task}
Each MR training instance contains at least the video name \texttt{<video>}, a query phrase \texttt{\{query\}}, and the ground-truth \texttt{\{start\}} and \texttt{\{end\}} timestamps. 
We construct diverse instruction templates and randomly sample one for each example to enhance linguistic variability. 
The prompt templates and expected output format are summarized in Table~\ref{tab:mr_instructions}. Examples before and after the conversion are as follows:

\begin{table*}[t]
\centering
\caption{Prompt templates and output format for the MR task.}
\label{tab:mr_instructions}
\footnotesize
\begin{tabularx}{\linewidth}{@{}lX@{}}
\toprule
\textbf{Type} & \textbf{Content} \\
\midrule
\textbf{Prompt Templates} & 
\texttt{<video>At what point in the video did the following events occur: \{query\}? Output the start and end timestamps.} \\
& \texttt{<video>What is the location of the moment: \{query\}?} \\
& \texttt{<video>Find when the following event happens in the video: \{query\}. Give me the start and end times.} \\
& \texttt{<video>Please indicate the start and end timestamps for the event: \{query\}.} \\
& \texttt{<video>Please predict start and end time of the following moment: \{query\}.} \\
& \texttt{<video>During which time interval does this happen in the video: \{query\}?} \\
& \texttt{<video>Locate the moment in the video where this occurs: \{query\}. Provide start and end times.} \\
& \texttt{<video>For the video, when does this event take place: \{query\}? Answer with start and end timestamps.} \\
& \texttt{<video>I want to know the start and end times of the following event in the video: \{query\}.} \\
& \texttt{<video>Could you tell me from what time to what time this happens: \{query\}?} \\
& \texttt{<video>Can you tell me the time window of this event: \{query\}?} \\
& \texttt{<video>Please find the timestamps that mark the occurrence of this event: \{query\}.} \\
& \texttt{<video>Identify the start and end of the following event in the video: \{query\}.} \\
\midrule
\textbf{Expected Output} & \texttt{from \{start\}s to \{end\}s} \\
\bottomrule
\end{tabularx}
\end{table*}

\noindent\textbf{Example 1 (Charades-STA).}

Original annotation:
\begin{lstlisting}[basicstyle=\ttfamily\footnotesize,breaklines=true,breakatwhitespace=true]
Y6R7T 20.8 30.0##person start playing on their phone.
\end{lstlisting}

Reformatted instance:
\begin{lstlisting}[basicstyle=\ttfamily\footnotesize,breaklines=true,breakatwhitespace=true]
{
  "messages": [
    {"role":"user",
     "content":"<video>At what point in the video did the following events occur: person start playing on their phone. Output the start and end timestamps."},

    {"role":"assistant",
     "content":"from 20.8s to 30.0s"}
  ],
  "videos":["Y6R7T.mp4"]
}
\end{lstlisting}

\noindent\textbf{Example 2 (ActivityNet-Captions).}

Original annotation (compact):
\begin{lstlisting}[basicstyle=\ttfamily\footnotesize,breaklines=true,breakatwhitespace=true]
{"video_id":"v_nwznKOuZM7w",
 "timestamps":[[0,15.51],[11.39,31.65]],
 "sentences":[
   "Two people are seen moving around a kitchen quickly performing various tasks and sitting down.",
   "They then wax down a ski in the kitchen while continuing to move around."
 ]}
\end{lstlisting}

Reformatted instances:
\begin{lstlisting}[basicstyle=\ttfamily\footnotesize,breaklines=true,breakatwhitespace=true]
{"messages":[
  {"role":"user",
   "content":"<video>What is the location of the moment: Two people are seen moving around a kitchen quickly performing various tasks and sitting down."},
  {"role":"assistant","content":"from 0s to 15.51s"}
 ],
 "videos":["v_nwznKOuZM7w.mp4"]},

{"messages":[
  {"role":"user",
   "content":"<video>Can you tell me the time window of this event: They then wax down a ski in the kitchen while continuing to move around."},
  {"role":"assistant","content":"from 11.39s to 31.65s"}
 ],
 "videos":["v_nwznKOuZM7w.mp4"]}
\end{lstlisting}

\subsection{Highlight Detection Task}
Each HD training instance contains at least the video name, a query phrase, relevant temporal segments, and their saliency scores. 
The prompt templates and output format are summarized in Table~\ref{tab:hl_instructions}. Specific examples are as follows:

\begin{table*}[t]
\centering
\caption{Prompt templates and output format for HD task.}
\label{tab:hl_instructions}
\footnotesize
\begin{tabularx}{\linewidth}{@{}lX@{}}
\toprule
\textbf{Type} & \textbf{Content} \\
\midrule
\textbf{Prompt Templates} & 
\texttt{<video>Please highlight the most exciting parts related to: \{query\}.} \\
& \texttt{<video>Find the most relevant or important moments for: \{query\}.} \\
& \texttt{<video>Which moments in the video best reflect: \{query\}?} \\
& \texttt{<video>Highlight the key segments that correspond to: \{query\}.} \\
& \texttt{<video>Show the most interesting clips about: \{query\}.} \\
& \texttt{<video>What are the highlight moments for: \{query\}?} \\
& \texttt{<video>Mark the time intervals that are most significant for: \{query\}.} \\
\midrule
\textbf{Expected Output} & 
\texttt{The highlights are: very important from \{start\}s to \{end\}s, ...; important from \{start\}s to \{end\}s, ...; less important from \{start\}s to \{end\}s, ...} \\
\bottomrule
\end{tabularx}
\end{table*}

\noindent\textbf{Example (QVHighlights).}

Original annotation (compact):
\begin{lstlisting}[basicstyle=\ttfamily\footnotesize,breaklines=true,breakatwhitespace=true]
{"qid":10016,
 "query":"Man in baseball cap eats before doing his interview.",
 "vid":"j7rJstUseKg_210.0_360.0",
 "relevant_windows":[[96,114]],
 "saliency_scores":[
   [2,3,3],[4,3,2],[2,3,1],[2,3,0],
   [2,3,3],[2,3,2],[2,3,1],[2,3,0],[1,3,3]
 ]}
\end{lstlisting}

Reformatted instance:
\begin{lstlisting}[basicstyle=\ttfamily\footnotesize,breaklines=true,breakatwhitespace=true]
{
  "messages": [
    {"role":"user",
     "content":"<video>Please highlight the most exciting parts related to: Man in baseball cap eats before doing his interview."},

    {"role":"assistant",
     "content":"The highlights are: important from 96.0s to 98.0s, 98.0s to 100.0s, 104.0s to 106.0s; less important from 100.0s to 102.0s, 102.0s to 104.0s, 106.0s to 108.0s, 108.0s to 110.0s, 110.0s to 112.0s, 112.0s to 114.0s"}
  ],

  "videos":["j7rJstUseKg_210.0_360.0.mp4"]
}
\end{lstlisting}

Based on the above methods, Grounding-FT reformulates heterogeneous VTG annotations into unified, instruction-response pairs. 
The diversity of prompt phrasing and conversational structure better aligns the dataset with large language model fine-tuning paradigms, leading to improved robustness and generalization.

\section{Discussion on Early vs. Multi-Stage Token Sampling}
In this work, VTS performs query-guided token sampling before multimodal fusion.
We adopt this design because it is simple, efficient, and well aligned with the VTG setting: early suppression of query-irrelevant visual content helps reduce noise before constructing the joint representation.
This also makes the sampling behavior more interpretable, since token relevance is estimated directly from the text query and visual features prior to deeper cross-modal interactions.
At the same time, we acknowledge that later-layer or multi-stage sampling could leverage richer multimodal semantics and potentially improve token selection further.
Such designs may offer a different trade-off between efficiency, interpretability, and representational power.
We view this as an interesting direction for future work.

\section{Additional Qualitative Analysis}

To complement the qualitative study, we provide additional visualization examples for both GroundVTS-Q and GroundVTS-I.
All examples follow the same visualization format as Figure~5 in the main paper, where the bottom curve denotes the normalized token density produced by the VTS module, with higher peaks indicating segments the model regards as more relevant to the query.

\begin{figure*}
  \centering
  \begin{subfigure}{\linewidth}
    \includegraphics[width=\linewidth]{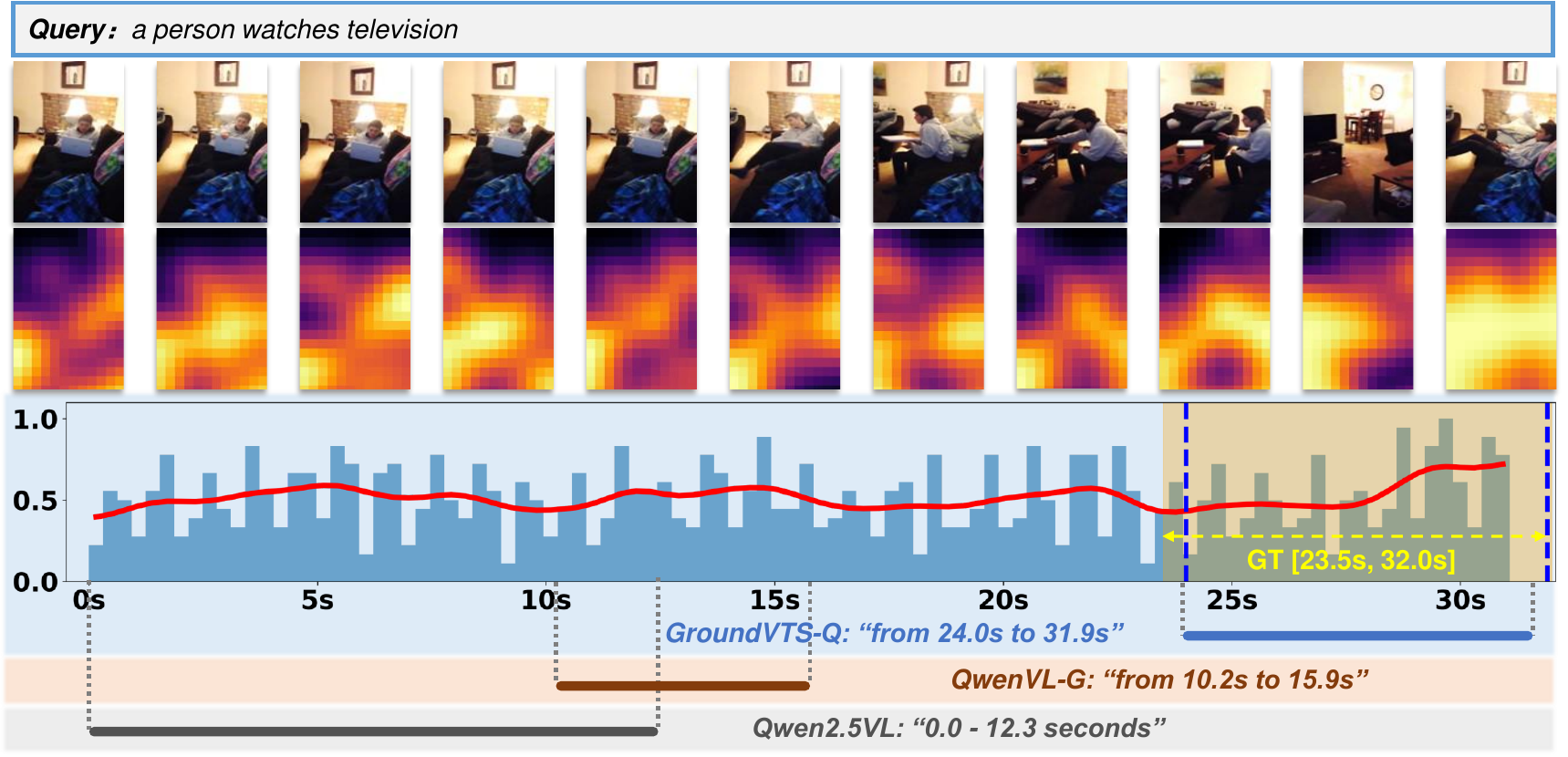}
    \caption{}
    \label{fig:suppl_qwen1}
  \end{subfigure}
  \begin{subfigure}{\linewidth}
    \includegraphics[width=\linewidth]{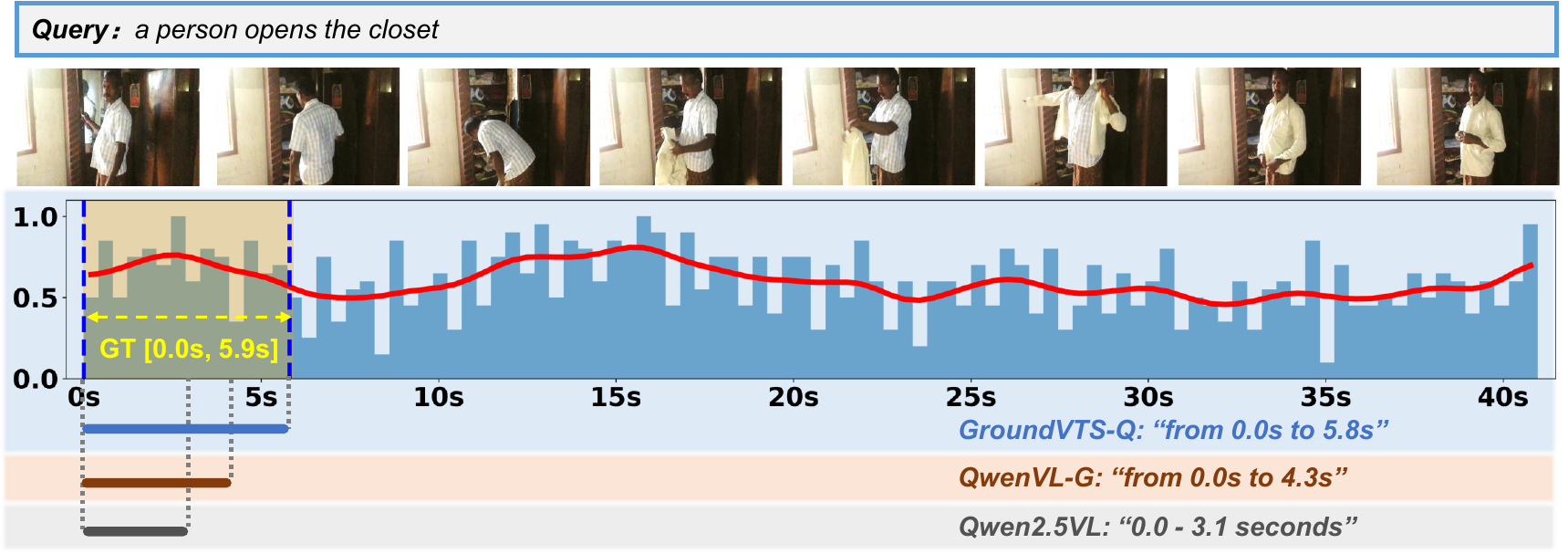}
    \caption{}
    \label{fig:suppl_qwen2}
  \end{subfigure}
  \caption{Additional qualitative comparison between GroundVTS-Q and its base models. Example (a) additionally illustrates spatial token retention maps, which correspond to spatial token selections.}
  \label{fig:suppl_qwen}
\end{figure*}

Across these cases, GroundVTS consistently exhibits highly accurate temporal localization. In Figure~\ref{fig:suppl_qwen}\subref{fig:suppl_qwen1} (GT: 23.5--32.0 s), GroundVTS-Q predicts 24.0--31.9 s, aligning almost perfectly with the ground-truth, whereas QwenVL-G and the pretrained Qwen2.5VL shift the interval far earlier and fail to localize the correct moment. 
A similar trend appears in Figure~\ref{fig:suppl_qwen}\subref{fig:suppl_qwen2} (GT: 0.0--5.9 s), where GroundVTS-Q outputs 0.0--5.8 s with near-exact precision, while both baselines truncate or deviate from the target boundary.

\begin{figure*}
  \centering
  \begin{subfigure}{\linewidth}
    \includegraphics[width=\linewidth]{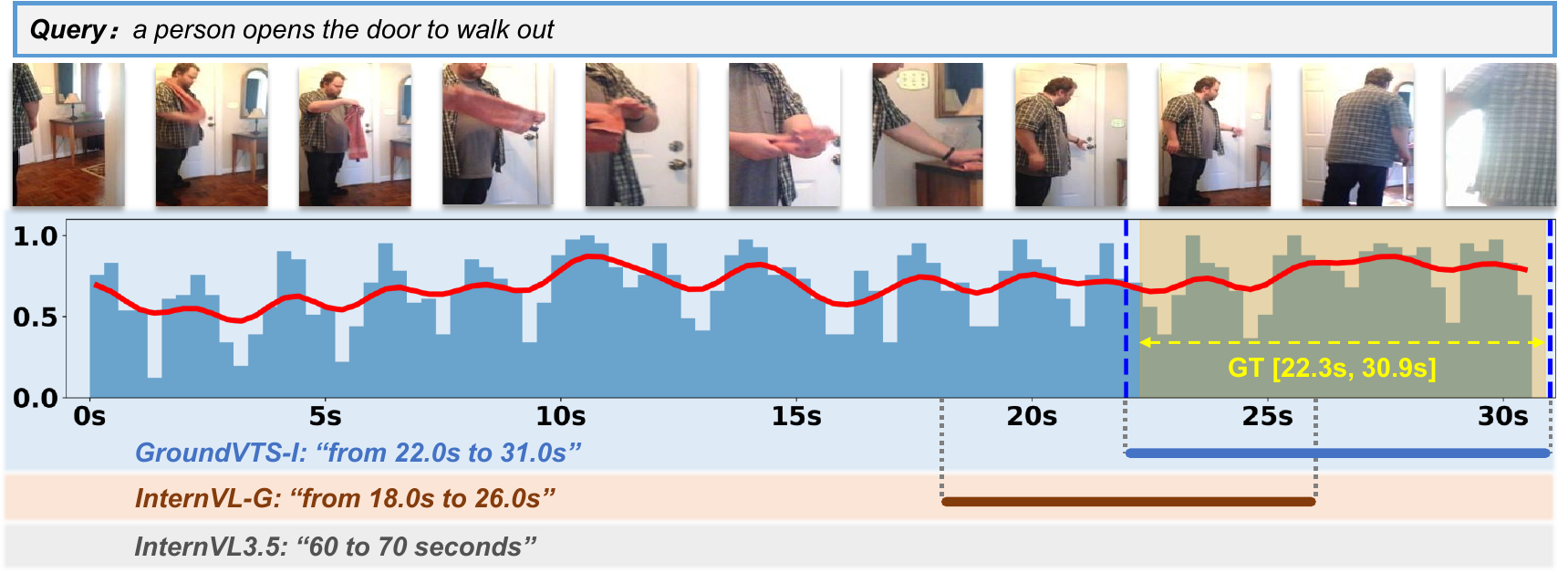}
    \caption{}
    \label{fig:suppl_intern1}
  \end{subfigure}
  \begin{subfigure}{\linewidth}
    \includegraphics[width=\linewidth]{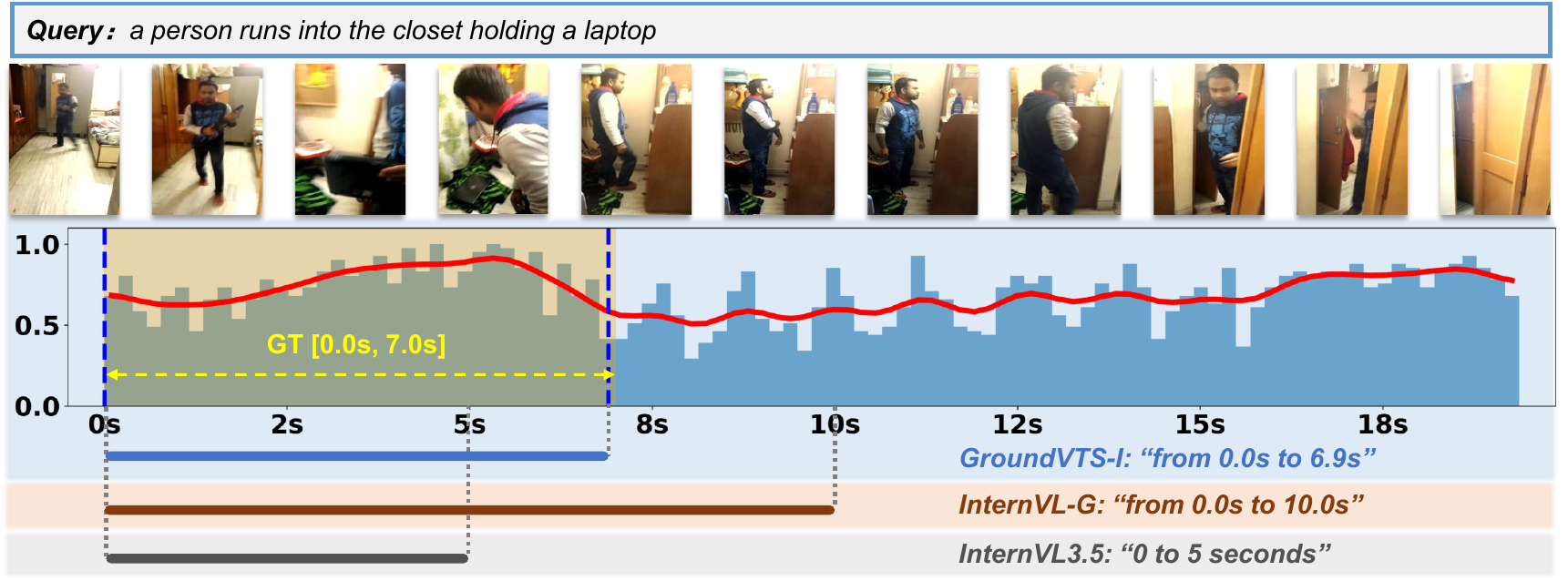}
    \caption{}
    \label{fig:suppl_intern2}
  \end{subfigure}
  \caption{Additional qualitative comparison between GroundVTS-I and its base models.}
  \label{fig:suppl_intern}
\end{figure*}

The InternVL-based examples exhibit the same trend. In Figure~\ref{fig:suppl_intern}\subref{fig:suppl_intern1} (GT: 22.3--30.9 s), GroundVTS-I produces a tightly aligned prediction of 22.0--31.0 s, whereas InternVL-G shortens and shifts the interval (18.0--26.0 s), and the base InternVL3.5 mislocalizes the event to a distant region.
In Figure~\ref{fig:suppl_intern}\subref{fig:suppl_intern2} (GT: 0.0--7.0 s), GroundVTS-I again matches the ground-truth boundaries accurately (0.0--6.9 s), while the baseline models either overextend the span or capture only a partial portion of the event.


These visualizations reveal three consistent advantages of GroundVTS.
First, its temporal predictions are markedly more accurate and better aligned with annotated spans, regardless of model backbone or event duration.
Second, its predicted spans consistently fall within the regions where the sampled token density reaches (local) maxima. In every example, the model’s final prediction aligns with the peaks of the VTS density curve, indicating that GroundVTS relies on the most informative temporal segments identified by the sampling module.
Third, the token allocation patterns produced by VTS are adaptive across different scenarios. In some cases (as illustrated in Figure~5 of the main paper), the density distribution forms sharp peaks with strong contrasts between attended and suppressed segments, typically corresponding to short or well-isolated grounding moments. In other cases, the differences between peaks and valleys are more moderate; nevertheless, the VTS curve still places a clear relative emphasis on the correct temporal region. These variations indicate that VTS does not rely on a fixed sparsity pattern but adjusts its sampling behavior according to the temporal structure of each video-query pair.

In addition, Figure~\ref{fig:suppl_qwen}\subref{fig:suppl_qwen1} visualizes the spatial distribution of visual tokens selected by VTS. It can be observed that VTS mainly focuses on the middle and lower regions of the frames, which correspond to areas around human activities. When the action of watching TV occurs at the end of the video, the VTS module attends to most regions of the frame.

Overall, by concentrating tokens at the most semantically relevant moments while downweighting redundant frames, VTS enables GroundVTS to encode fine-grained temporal cues more effectively, leading to significantly sharper and more accurate temporal boundaries.

\end{document}